\documentclass[
  equalmargins,
  12pt,
  openany,
  onehalfspacing,
]{ut-thesis}
\usepackage[colorlinks]{hyperref} 
\usepackage[backend=biber, sorting=none, style=ieee]{biblatex} 
\usepackage{graphicx} 
\usepackage{booktabs} 
\usepackage{amsmath}
\usepackage{amsthm}
\usepackage{amsfonts}
\usepackage{tensor}

\usepackage{algorithm}
\usepackage{algpseudocode}

\usepackage[table,xcdraw,dvipsnames]{xcolor}
\usepackage{subfigure}

\usepackage{times}
\usepackage{epsfig}
\usepackage{amsmath}
\usepackage{mathtools,amsfonts,bm}

\usepackage{mhchem}
\usepackage{lipsum}

\usepackage{float}
\usepackage{caption}

\usepackage{subfigure}
\usepackage{multirow}
\usepackage{booktabs}

\DeclareMathAlphabet{\mathsfit}{\encodingdefault}{\sfdefault}{m}{sl}
\SetMathAlphabet{\mathsfit}{bold}{\encodingdefault}{\sfdefault}{bx}{n}
\newcommand{\tens}[1]{\bm{\mathsfit{#1}}}
\def\tW{{\tens{W}}}

\definecolor{ao}{rgb}{0.0, 0.5, 0.0}

\DeclareMathOperator*{\argmin}{arg\,min}

\newcommand{\etens}[1]{\mathsfit{#1}}

\def\etF{{\etens{F}}}

\newtheorem{definition}{Definition}[section]

\theoremstyle{remark}

\author{Andre Fu}
\title{Leveraging Learning Metrics for Improved Federated Learning}
\degree{Bachelor's of Applied Science}
\department{Engineering Science}
\gradyear{2022}
\begin{document}
  \frontmatter
    \maketitle
    \begin{abstract}
      Currently in the federated setting, no learning schemes leverage the emerging research of explainable artificial intelligence (XAI) in particular the novel learning metrics that help determine how well a model is learning. One of these novel learning metrics is termed `Effective Rank' (ER) which measures the Shannon Entropy of the singular values of a matrix, thus enabling a metric determining how well a layer is mapping. By joining federated learning and the learning metric, effective rank, this work will \textbf{(1)} give the first federated learning metric aggregation method \textbf{(2)} show that effective rank is well-suited to federated problems by out-performing baseline Federated Averaging \cite{konevcny2016federated} and \textbf{(3)} develop a novel weight-aggregation scheme relying on effective rank.
\begin{center}
\href{https://github.com/andre-fu/jakaa}{\textcolor{blue}{\texttt{https://github.com/andre-fu/jakaa}}}
\end{center}

    \end{abstract}
    \tableofcontents
    \listoffigures
  \mainmatter
    \chapter{Introduction }\label{ch:intro} 
The traditional deep-learning (DL) paradigm relies on a central server, where data collected from external sources (mobile phones, laptops, etc) are sent to the server. Massive amounts of data transfer occur between these edge devices and the server, representing a serious communication overhead. In 2016, a new paradigm was proposed by Google \cite{konevcny2016federated}, termed Federated Learning (FL) where edge devices download copies of the model from the central server and train locally. Trained weights are sent back to the central server to be averaged, updating the global model. This novel federated learning system allowed for a completely new method of training deep-learning models, distributing the load of training while providing privacy-preserving features that mitigate modern privacy concerns. \\

\section{Motivation}
As deep learning evolves, the requirements we, as a society, expect of DL increases in parallel. These expectations are built on the behemoth amount of data that exists in both structured and unstructured forms in the real world. Unfortunately, these larges swaths of data that could be leveraged for training DL models, is hidden behind `data islands' where the organizations holding this data tend to be highly protective. A canonical example within the FL community are the highly valuable datasets within hospitals \cite{li2021flsurvey}. Hospitals protect patient privacy through internal policies meaning they often are unable to share patient data without permission. The phenomena of `data islands' permeates multiple verticals such as government, medicine, finance, retail and supply chains. As such, leveraging this large swath of data while maintaining the private nature of the data would be mutually beneficially to all parties involved. \\

\section{Problem Formulation}
\subsection{Federated Learning}\label{ch:fedlearn}
Within federated learning, two paradigms emerge, firstly the ``cross-device'' learning which outlines the initial use-case: for multiple edge devices to collaboratively train networks \cite{konevcny2016federated}. Secondly, as the field progressed, ``cross-silo'' use-cases emerged, such as the hospital scenario, where trusted organizations can serve as reliable clients. In light of these advances, a new definition of federated learning was developed \cite{kairouz2019advances}:

\begin{definition}[Federated Learning]\label{deffl}
Federated learning is a machine learning setting where multiple entities (clients) collaborate in solving a machine learning problem, under the coordination of a central server or service provider. Each client’s raw data is stored locally and not exchanged or transferred; instead, focused updates intended for immediate aggregation are used to achieve the learning objective.
\end{definition}

As per \autoref{deffl}, the question of coordination, and aggregation of client updates naturally arises. Multiple methods have been proposed based on the initial federated stochastic gradient descent (FedSGD) algorithm \cite{li2020preserving}, which extends stochastic gradient descent (SGD) to a federated setting. One of the most simple and popular techniques is termed federated averaging where clients get the global model shared by the central server, training occurs, and a model is returned after an epoch. The averaging of model weights is done via simple averaging. 
\begin{figure}[htp]
    \centering
    \includegraphics[width=\textwidth]{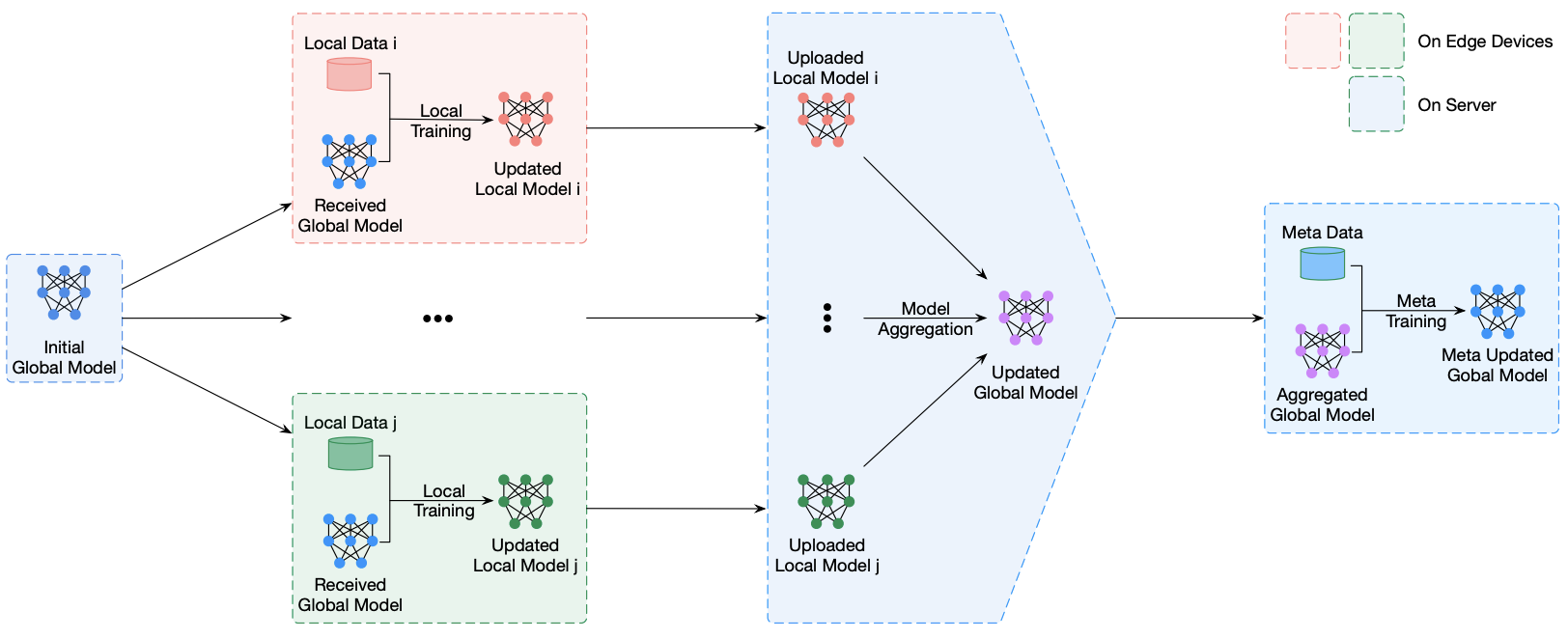}
    \caption{Federated Averaging where model parameters are shared in consecutive actions by the server-client model. \cite{konevcny2016federated}}
    \label{fig:fedavg}
\end{figure}

Building on this work, more interesting re-weightings of the client weights have taken place, such re-weighting the client parameters according to the inverse of the un-centered variance of the SGD operation at every client \cite{reyes2021precision}. Implicitly this means that models that do not have as many data points to train, will have higher variance, and thus lower weight in model aggregation. Currently, the FL community has not designated one aggregation scheme as the `go-to' method for knowledge aggregation. In this space, exploration in optimizing aggregation schemes is still an open-question allowing for novel model parameter aggregation techniques to be evaluated. \\

\subsection{Learning Metrics}\label{ch:learnmetrics}
As the Artificial Intelligence (AI) community begins more earnest research into explainable AI (XAI), a theme of developing explainable metrics to give human intuition \cite{xai} around the black-box models has risen. A subset of this phenomena is the idea of developing `learning' metrics, to probe the internal layers of a neural network in an effort to analyze how well the model is learning. By assessing knowledge aggregation metrics during training, intuition around model function, mapping space and feature representations can be gained and thus allowing humans to in turn, develop better DL models. \\

In light of this, \cite{jaegerman21genprob} has shown multiple novel learning metrics \textit{stable rank, condition number, effective rank} and \textit{quality measure}. These learning metrics probe the intermediate convolutional layers to better understand the mapping space between any two layers.

\section{Objectives}
The primary purpose of this thesis will be to marry the ideas developed in \autoref{ch:learnmetrics}, `learning metrics' and  \autoref{ch:fedlearn}, `federated learning' to develop a novel re-weighting scheme where model weights will be aggregated in proportion to the Effective Rank \cite{jaegerman21genprob}.
    \chapter{Literature Review}
This literature review assumes some basic knowledge of statistics, vector calculus and distributed systems. It will then focus on the nature of machine learning and deep learning from a theoretical then practical perspective, serving as the basis for the  mathematical explanations formed in later chapters.
\section{Federated Learning Aggregation Methods}
\textbf{Optimization Problem Formulation}\\
Due to the nascent nature of federated learning, multiple weight aggregation methods have been proposed, but they all follow the original problem formulation. In \cite{konevcny2016federated} the optimization scheme to create communication efficient is proposed as:
\begin{equation}
    \min_{\textit{w} \in \mathbb{R}^d} f(w) \text{ \quad where \quad} f(w) = \frac{1}{n} \sum_{i=1}^{n} f_i (w) 
    \label{eq:optform}
\end{equation}
For machine learning problems $f_i (w) = \ell(x_i, y_i; w) $, the loss of prediction sample $(x_i, y_i)$ using weight $w$. With $K$ clients, on selection $k \in K$ are chosen to participate in the model update, whose data $\mathcal{D}_k$ is the data on the $k$th client, where $n_k = |\mathcal{D}_k|$ and $n = \sum n_k$. In this way, we rewrite \autoref{eq:optform} as:
\begin{equation}
    f(w) = \sum_{i = 1}^{n_k} \frac{n_k}{n} F_k (w) \text{ \quad where \quad} F_k(w) = \frac{1}{n_k} \sum_{i \in \mathcal{D}_k} f_i(w)
    \label{eq:optform2}
\end{equation}
\textbf{Stochastic Gradient Descent in a Federated Setting}\\
Gradient updates in traditional machine \& deep learning take place with weight updates using the stochastic gradient descent (SGD) algorithm as follows:
\begin{equation}
    w^{(t+1)} \leftarrow w^{(t)} + \eta_t \nabla f_i(w)
\end{equation}
For some step size, $\eta$, predefined by iteration $t$. 
In a federated setting, the central server aggregates the weights via 
\begin{equation}
    w^{(t+1)} \leftarrow w^{(t)} + \eta_t \sum_{i = 1}^{K} \frac{n_k}{n} g_k  \text{ \quad where \quad} g_k = \nabla F_k(w_t)
    \label{eq:grad1}
\end{equation}
It follows then that $\sum_{i = 1}^{K} \frac{n_k}{n} g_k = \nabla f_i(w) $. We can then rewrite \autoref{eq:grad1} into a simpler formulation:
\begin{align}
    w^k_{t+1} \leftarrow  w^k_{t} + \eta \cdot g_k \qquad \text{Client Update} \label{eq:client_fsgd}\\
    w_{t+1} \leftarrow \sum_{i = 1}^{K} \frac{n_k}{n}w^k_{t+1} \label{eq:server_fsgd} \qquad \text{Server Update}
\end{align}

Since this is the case the local models can update multiple more steps (termed rounds, $E$ - traditionally the minibatch update) before an update occurs on the server.

\subsection{Precision Weighted Federated Learning}
In \cite{reyes2021precision} the authors note that the data heterogeneity may be overestimated in \cite{konevcny2016federated}, and in doing so they mistake the variance within the data. The authors, then propose that leveraging the intra-variability to improve performance through evaluation of the dataset variance in the gradient update. \\

In \cite{konevcny2016federated}, it is noted that for independent and identically distributed (i.i.d) data the variance $\mathbb{E}_{\mathcal{D}_k}[F_k(w)] = f(w)$, but when the data is heterogeneous, this assumption does not hold. Therefore by replacing the weighted averaging with the variance of the maximum likelihood estimator (MLE) in \autoref{eq:server_fsgd}.
\begin{equation}
     w_{t+1} \leftarrow \sum_{i = 1}^{K} \frac{(v_{t+1}^i)^{-1}}{\sum_{i=1}^{K} (v_{t+1}^i)^{-1}}w^k_{t+1} 
\end{equation}
Where $v_{t+1}^k$ denotes the variance of the MLE of $w$, which intuitively considers the model's uncertainty in choosing $w$. 

\subsection{Unbiased Gradient Accumulation}
The authors of \cite{yao2019federated} show that as $E$ increases, the difference between the ideal $g_t$ computed at every iterative update and the $g_t^{k(i)}$ for the $i$th of $E$ steps grows larger. They denote this gap the gradient bias, which initially is small but accumulates for large $E$ as $t$ continues. Furthermore, they also note the differing optimization objectives within a federated setting when choosing a subset of the clients to represent the whole dataset. By showing that the target distribution of data $\mathcal{D}$ may not match the chosen subset from the clients (denoted $\mathcal{D}_{S_t}$) the assumption works well for cross-silo situations but in the cross-device paradigm, the edge compute may not be representative, or able to sustain training and thusly have differing objectives from the true optimization objective.\\

In order circumvent the difference accumulation in the gradients, the computation in \autoref{eq:server_fsgd} of $w^{k(i)}_{t+1}$ can be replaced directly with $w_t$ which violates the client update rule \& requires more than necessary compute. On client for the $i$th step of the $t$th round, the update looks like: 
\begin{equation}
    w^{k(i)}_{t} \leftarrow  w^{k(i-1)}_{t} + \eta g_t^{k(i)}
\end{equation}
Therefore, to minimize compute and allow client updates, the authors in \cite{yao2019federated} propose instead of computing the numerical value of $ w^{k(i)}_{t}$, to instead keep the relationship between $ w^{k(i)}_{t}$ and $ w^{k(i-1)}_{t}$, termed `keep-trace' of gradients. Then, on the final $E$th epoch, evaluate the aggregated gradients. \\

To address the distribution mismatch, the authors propose a meta dataset $\mathcal{D}_{meta}$ which would evaluate the aggregated weights against the meta-dataset, and perform a meta-weight update that would attempt to `re-align' the underlying dataset selection $\mathcal{D}_{S_t}$ to the true dataset distribution $\mathcal{D}$.

\subsection{Federated Proximal Learning}
To address the heterogeneity concerns from \cite{konevcny2016federated}, the authors from \cite{li2020federated} propose including straggling clients, which may adversely affect performance. To mitigate the performance degradation, the authors include a proximal term to improve stability allowing the server to numerically associate the statistical heterogeneity within clients. They achieve this goal through relaxation of the exactness of the local objectives' optimization. By solving the local objectives inexactly in some $\gamma$-bound, where the number of local epochs, $E$ on device can differ. The amount of $\gamma$-inexactness is calculated per device, per round $t$ allowing the central server to include this bound in the aggregation. \\

\begin{definition}[$\gamma_k^t$-inexactness]\label{def:gminexact}
For function $h_k(w;w_t)  = F_k(w) + \frac{\mu}{2}\Vert w - w_t\Vert^2$ and $\gamma \in [0, 1]$ we say $w^*$ is a $\gamma_k^t$-inexact solution of $\min_{w} h_k$ if $\Vert \nabla h_k(w^*; w_t) \Vert \leq \gamma_k^t \Vert \nabla h_k(w_t; w_t) \Vert$, where $ \nabla h_k(w; w_t) = \nabla F_k(w) + \mu(w-w_t)$
\end{definition}

By comparing the variation between the learned $w$ and the original  $w_t$ sent from the server, the optimization chooses a $w_k^{t+1} \approx \argmin_w h_k(w;w_t) = \argmin_w F_k(w) + \frac{\mu}{2}\Vert w - w_t\Vert^2$. This effectively regulates the local choice $w$ from straying too far from the global optimization. 

\section{Learning Metrics}
\textbf{Convolutional Neural Networks}\\
Convolutional Neural Networks (CNNs) were originally developed in the landmark AlexNet paper \cite{krizhevsky2012imagenet} where Graphical Processing Unit (GPU) was used to parallelize the matrix operations required for sliding kernels over an input image. Since then CNNs have developed significantly more, with Residual Networks \cite{he2016deep} showing the topology of a CNN can be interpreted as a Directed Acyclic Graph (DAG) with skip connections. From this, multiple works have attempted to explain how CNNs learn and what knowledge is contained within the feature extraction tensors. 

\subsection{Feature Extraction through Tensor Decomposition}
CNN learned weigths are a 4-dimensional tensor with shape $k\times k \times N_3 \times N_4$ where $N_3 = C_{in}$ is the input channels and $ N_4 = C_{out}$ is the output channels and $k\times k$ denotes the kernel size. During convolution, the output feature map is convoluted in the following fashion: $\etF^{O}_{:,:,i_4} = \sum_{i_3=1}^{N_3}\etF^{I}_{:,:,i_3}\ast\tW_{:,:,i_3,i_4}$, where ${\etF}^{O}\in\mathbb{R}^{H \times W \times N_{4}}$. 

Since the convolutional weight $\tW$ acts as a learned encoder for feature extraction, Fu \textit{et al} \cite{fu2021conet} was able to unfold this 4D tensor along a dimension, $d$ i.e $\tW [\texttt{4D-tensor}] \rightarrow \tW_d [\texttt{2D-matrix}]$. Now that the tensor is in matrix-form, we can utilize the tools within the 2D domain to asses the weights. Specifically, \cite{fu2021conet} we can extract the noisy initialization from the more valuable learnt encodings, though a Variational Bayes Matrix Factorization (VBMF) \cite{nakajima2013global}. Effectively, \cite{fu2021conet} does the following : $\tW_d = \widehat{\tW_d} + E_{noise}$, where $ \widehat{\tW_d}$ is a low-rank Singular Value Decomposition (SVD). This method takes scales with arbitrary size of a matrix, and is computationally fast on CPU, freeing GPU resources for training.

\begin{figure}[htp]
    \centering
    \includegraphics[width=\textwidth]{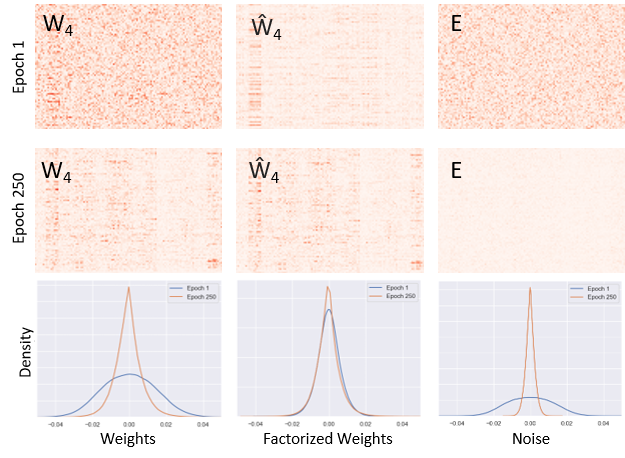}
    \caption{Low Rank Factorization for ResNet, showing noise reduction during training and learned encoding knowledge}
    \label{fig:lowrankdist}
\end{figure}

\subsection{Effective Rank}\label{sec:er}
Building on \cite{fu2021conet}, the work proposed by \cite{jaegerman21genprob} shows that multiple derivative explainability, learning \& quality metrics can be extracted from the $\widehat{\tW_d}$ representation. While the authors test multiple metrics, the effective rank $ER$ had the lowest test-train generalization gap, while maintaining the closest correlation to the test-accuracy. In \cite{jaegerman21genprob}, the effective rank is a borrowed metric from \cite{roy2007effective}, representing the dimension of output space from the transformation operated by a deep layer, calculated using Shannon Entropy of the normalized values of a layer. \\

On a layer-by-layer basis Effective Rank, $ER$ is calculated as follows, where $n'$ denotes the rank of matrix $\tW$ and $\sigma_k (\tW)$ denotes the normalized singular values of Matrix $\tW$:
\begin{equation} \label{eqn:er}
    ER = \sum_{i=1}^{n'} \bar{\sigma_k}(\tW_i) \log \bar{\sigma_k}(\tW_i)
\end{equation}
In order to evaluate an entire layer we utilize the aggregation below:
\begin{equation}\label{eqn:modeler}
    Q_{ER} = \log \Bigg( \sqrt{\sum_{i=1}^d ER(\tW_i)^2 } /d \Bigg)
\end{equation}

Since  \autoref{eqn:er} is the Shannon entropy of the singular values of $\tW$, it measures the amount of entropy of the singular values distribution. As weight matrices $\tW$ maps input and output feature representations, the singular values represent the `axes' of this transformation. As such measuring the entropy of the scaling singular values $\bar\sigma_k$ intuitively represents the amount of `information mapping' between two feature representations.
 

    \chapter{Methods \& Algorithm Development}
In the traditional framework of federated optimization, the central server will aggregate weights from the client models. This implementation is termed Federated Effective Rank (FedER). This section will first explain some intuition and explain the methodology, then explain the naive implementation that performed poorly and then go into a secondary improved version. 
\section{Intuition \& Methodology }

\begin{figure}[H]
    \centering
    \includegraphics[width=\textwidth]{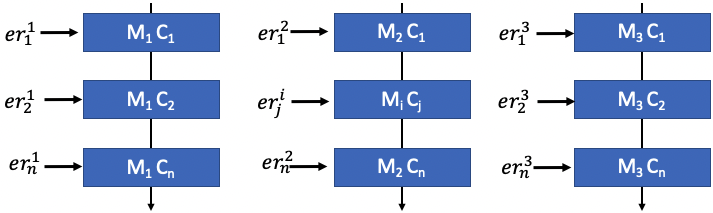}
    \caption{Effective Rank Measured Across Models; }
    \label{fig:er-across-models}
\end{figure}

Above in \autoref{fig:er-across-models} the general methodology is laid out. Each of the `arrows' with blue-blocks represents a model ($M_i$) where the blue blocks represent convolutional blocks/layers ($C_j$). The Effective Rank is calculated per layer as $er_j^i$ where $j$ refers to the convolutional block $C_j$ (may also be referred to as layer $l$) and $i$ refers to the model $M_i$ (also referred to client $k$). After calculating the effective rank for all the convolutional blocks, the weighted-average of the model's weights need to occur. This is done on a layer-by-layer basis so that every convolutional block's effective rank is weighted in proportion to that layer. As such \autoref{eq:alphweight} shows how the weighting $\alpha$ is calculated according to effective rank. Then $alpha$ will be used when averaging the model weights to aggregate information.

\begin{equation}\label{eq:alphweight}
    \alpha_l^k \leftarrow \frac{er_l^k}{\sum_{k} er_l^k} 
\end{equation}
\\
Below in Algorithm \autoref{alg:clientupdate} and Algorithm \autoref{alg:effectiverank} shows the two common algorithms that do not change throughout this implementation. In effect Algorithm \autoref{alg:clientupdate} refers to \autoref{eq:grad1} while Algorithm \autoref{alg:effectiverank} refers to \autoref{eqn:er}.

\begin{algorithm}
\caption{Common Client Update}\label{alg:clientupdate}
\begin{algorithmic}[1]
\Function{ClientUpdate}{$w_t^k$}
    \For {epoch $e = 1 ... E$ }
        \For {batch $b \in \mathcal{B}$}
            \State $w_t^k \leftarrow w_t^k - \eta \nabla \ell(w_t^k, b)$
        \EndFor
    \EndFor
    \State \textbf{return} $w_t^k$
\EndFunction
\end{algorithmic}
\end{algorithm}

\begin{algorithm}
\caption{Effective Rank}\label{alg:effectiverank}
\begin{algorithmic}[1]
\Function{EffectiveRank}{$\tW$}
    \State $U, S, V \leftarrow$ \texttt{EVBMF}($\tW$) 
    \State \textbf{return} \texttt{ShannonEntopy}(S)
\EndFunction
\end{algorithmic}
\end{algorithm}

\newpage
\section{Initial Weight Aggregation}\label{sec:initmethod}
Simple weight aggregation though effective rank calculation can take place by iterating through the layers of a Convolutional Neural Network (CNN) and calculating the effective rank. 

\begin{algorithm}
\caption{Initial Algorithm development of FedER}\label{alg:init}
\begin{algorithmic}[1]
\Procedure{Server Execution}{}
\State Initialize: $w_0$, 
\For {round $t = 1, 2 ... T$}
    \For {client $k \in K$ in parallel}
        \State $w_{t}^{k} \leftarrow$ \texttt{ClientUpdate}$(w_{t}^{k})$ \Comment{Alg \autoref{alg:clientupdate}}
    \EndFor
    \State Initialize Effective Rank Aggregator: $A_l$
    \For {layer $l = 1 ... L$ in $w_{t}^{k} = [\tW_1^{k} ... \tW_l^{k}...\tW_L^{k}]$ }
        \If {dim($\tW_l^{k}$) = 4} \Comment{if convolutional layer}
            \State $er_{l}^{k} \leftarrow $ \texttt{EffectiveRank}($\tW_l^{k}$) \Comment{ Alg \autoref{alg:effectiverank}}
            \State $A_l \leftarrow A_l + er_{l}^{k}$
        \EndIf
    \EndFor
    \State \textit{// Begin Averaging According to Effective Rank}
    \For {layer $l = 1 ... L$ in $w_{t}^{k} = [\tW_1^{k} ... \tW_l^{k}...\tW_L^{k}]$ }
        \If {dim($\tW_l^{k}$) = 4} \Comment{if convolutional layer}
            \State $er_{l}^{k} \leftarrow $ \texttt{EffectiveRank}($\tW_l^{k}$) \Comment{ Alg \autoref{alg:effectiverank}}
            \State $\alpha_l^k \leftarrow er_{l}^{k}/A_l$
            \State $\tW_l \leftarrow \sum_{k \in K} \alpha_l^k \cdot \tW_l^{k}$
        \EndIf
        \If {dim($\tW_l^{k}$) $\neq$ 4} \Comment{If non convolutional layer - simple averaging}
            \State $\tW_l \leftarrow \sum_{k \in K} (1/K) \cdot \tW_l^{k}$
        \EndIf
    \EndFor
\EndFor
\EndProcedure
\end{algorithmic}
\end{algorithm}

\subsection{Limitations}
This naive implementation essentially sets the `effective rank' of the non-convolutional weights to 1, thus relying on an implicit `mapping' of one. This hinders the model's ability to utilize the non-convolutional effective rank, or propagate the previous layer's effective rank to the non-convolutional blocks. \\

\section{Improved Effective Rank Weight Aggregation}
To improve on the approach outlined in \autoref{sec:initmethod}, two key changes are made. Firstly the pure effective rank is not considered, but rather the maximum between 1e-3 and the effective rank. Secondly, for non-convolutional blocks, the previous convolutional layer's effective rank is used.\\

Similar to Algorithm \autoref{alg:init}, the server first trains the clients then comes together to aggregate the results. A few key changes are made, in particular Line 10 where the maximum between 0.001 and the calculated effective rank was taken. When a model begins training, it is effectively a randomized mapping, which provides no value. Seen in \autoref{fig:lowrankdist} where the weights `uncover' (learn) the mapping during training. \\

\begin{algorithm}[H]
\caption{Improved Implementation of FedER}
\begin{algorithmic}[1]
\Procedure{Server Execution}{}
\State Initialize: $w_0$, 
\For {round $t = 1, 2 ... T$}
    \For {client $k \in K$ in parallel}
        \State $w_{t}^{k} \leftarrow$ \texttt{ClientUpdate}$(w_{t}^{k})$ \Comment{Alg \autoref{alg:clientupdate}}
    \EndFor
    \State Initialize Effective Rank Aggregator: $A_l$
    \For {layer $l = 1 ... L$ in $w_{t}^{k} = [\tW_1^{k} ... \tW_l^{k}...\tW_L^{k}]$ }
        \If {dim($\tW_l^{k}$) = 4} \Comment{if convolutional layer}
            \State $er_{l}^{k} \leftarrow $ max(1e-3, \texttt{EffectiveRank}($\tW_l^{k}$)) \Comment{ Alg \autoref{alg:effectiverank}}
            \State $A_l \leftarrow A_l + er_{l}^{k}$
        \EndIf
    \EndFor
    \State \textit{// Begin Averaging According to Effective Rank}
    \For {layer $l = 1 ... L$ in $w_{t}^{k} = [\tW_1^{k} ... \tW_l^{k}...\tW_L^{k}]$ }
        \State Initialize: $\alpha_p^k$ \Comment{The previous $p$ conv layer}
        \If {dim($\tW_l^{k}$) = 4} \Comment{if convolutional layer}
            \State $er_{l}^{k} \leftarrow $ \texttt{EffectiveRank}($\tW_l^{k}$) \Comment{ Alg \autoref{alg:effectiverank}}
            \State $\alpha_l^k \leftarrow er_{l}^{k}/A_l$
            \State $\alpha_p^k \leftarrow \alpha_l^k$ \Comment{Save the last convolutional block's $\alpha$}
            \State $\tW_l \leftarrow \sum_{k \in K} \alpha_l^k \cdot \tW_l^{k}$
        \EndIf
        \If {dim($\tW_l^{k}$) $\neq$ 4} 
           \State $\tW_l \leftarrow \sum_{k \in K} \alpha_p^k \cdot \tW_l^{k}$ \Comment{Use last convolutional ER for non-conv}
        \EndIf
    \EndFor
\EndFor
\EndProcedure
\end{algorithmic}
\end{algorithm}


\subsection{Limitations}
This implementation suffers from a few key limitations, firstly the effective rank calculation is done entirely through one a single process, and could be done in parallel to improve efficiency. Secondly, the weight aggregation could be vectorized instead of done in consecutive \texttt{for} loops. While technically the asymptotic nature of this implementation is still an $\mathcal{O}$(n) the multiple iterations means lost computational efficiency. Thirdly, this weight aggregation scheme suffers from a few-high level limitations. 

In particular, it would struggle to reconcile multiple different architectures. For example if one client was running the residual network with 18 layers (ResNet 18) and another running with 34 layers (ResNet 34) \cite{he2016deep} the layer-by-layer approach would fail. This could be resolved by calculating the model effective rank in \autoref{eqn:modeler}. Another limitation is around the asynchronous nature of distributed systems, where certain client could drop from training. In these cases, the clients could inhibit training since the central server expects all clients to be together or the client could become a free rider \cite{fraboni2021free}. A free rider is a client that does not contribute to training but receives the finalized trained model at the end, thus getting the model for `free'. In our specific case of cross-silo implementations this concern becomes less severe, but is still present with hostile clients.

    \chapter{Results \& Discussion}
In this section, the results from a series of experiments comparing the federated effective rank averaging method (FedER) to state-of-the-art methods and tested across 3 different optimizes with and without learning rate scheduler StepLR. \\

\textbf{Figure Description: }All figures in this section will have an accuracy and loss plot. The Accuracy title will have the Top-1 Test Accuracy immediately after the title. The Legend will contain numbers \{0, 1, 2, 3 \} denoting Client 0, Client 1, Client 2, Client 3. The pink line in the legend will denote the test-accuracy measured at the end of every round. 

\section{Optimizers}\label{sec:optimizers}
\textbf{Adam:} Adam \cite{kingma2014adam} is the most widely used optimizer in deep learning due to it's ability to easily outperform competing methods. Effectively Adam treats each parameter gradient as a distribution and computes the per-parameter-gradient mean and variance (then some correction terms), and leverages them to derice the  `coefficient of variance' (CV) which is used to update the parameters.\\

\textbf{AdamP:} AdamP \cite{heo2020adamp} is built on top of Adam, but instead of directly using the weights, they notice that for scale-invariant weights like Batch Norm layers (of which $>$80\% of ResNet is comprised) excessive growth occurs. Since the step-size of optimizers is inversely proportional to the norm of these weights, they may yield in sub-optimal weight selections for scale-invariant weights. As such, AdamP instead \textit{regularizes through projection} onto the tangent of the gradient. Thus, AdamP is able to leverage momentum while slowing growth for scale-invariant weights.\\

\textbf{RMSGD:} The Rank-Momentum SGD (RMSGD) method \cite{hosseini2022exploiting} is a novel method recently presented, that introduced learning metrics \textit{stable rank} measuring how well a layer encodes information and \textit{condition number} which measures stability with respect to perturbations \& noise. RMSGD is effectively a per-parameter learning rate scheduler that adaptively chooses a learning rate in order to guarantee a monotonic increase in stable rank. As such it can `force' better encoding of information per layer and thus close the generalization-gap.\\

\section{Computational Setup}
\textbf{Setup:} We evaluate `FedER' against Federated Averaging \cite{konevcny2016federated} as applied to Image Classification task CIFAR100 with cutout and using the Residual Network model with 18 layers (ResNet18)  \cite{he2016deep}. \\

\textbf{Hardware:} In the distributed setting, we use a single GPU (RTX2080Ti) per client, on 4 different clients meaning 4x RTX2080Ti. The server used an Intel Xeon Gold 6246 processor, and 256 gigabytes of RAM. No experiments were performed using multi-GPU parallelism. The model aggregation was conducted on CPU, and only training was conducted on GPU.\\

\textbf{Standardization:} Since optimizers could exhibit widely different epoch times, we limited training to 250 epochs using SGD. While there are many advantages to federated learning, in this work, for experimental results, we attempt to leverage the knowledge sharing nature to develop better overall models. Full configuration details are available below in \autoref{tab:config}\\

\begin{table}[htp]
\begin{center}
\begin{tabular}{c|c}
\textbf{Hyper-parameter}                                        & \textbf{Configuration}                                           \\ \hline
Batch Size                                                      & 128                                                              \\ \hline
\begin{tabular}[c]{@{}c@{}}Initial Learning\\ Rate\end{tabular} & 1e-3                                                             \\ \hline
Weight Decay                                                    & 1e-4                                                             \\ \hline
Loss Function                                                   & Cross-Entopy                                                     \\ \hline
Scheduler                                                       & \begin{tabular}[c]{@{}c@{}}(if applicable)\\ StepLR\end{tabular} \\
Scheduler Step Size                                             & 25                                                               \\
Scheduler Decay                                                 & 0.5                                                             
\end{tabular}
\end{center}
\caption{Hyper-parameter configurations for Experiments}\label{tab:config}
\end{table}
\newpage

\section{Experiments}
The following experiments are run on ResNet 18 with CIFAR100 in a non-federated setting, with Federated Averaging (FedAvg) and the novel Federated Effective Rank (FedER) method:
\begin{itemize}
    \item Adam without StepLR \vspace{-0.58cm}
    \item Adam with StepLR \vspace{-0.58cm}
    \item AdamP without StepLR \vspace{-0.58cm}
    \item AdamP with StepLR \vspace{-0.58cm}
    \item RMSGD without StepLR \vspace{-0.58cm}
\end{itemize}
\subsection{Non-Federated Baselines}
In this section, the same hyper-parameters ran in the federated setting, are run in the traditional setting to compare as a non-federated baseline. \\

\textbf{Figure Descriptions:} Similar to the federated setting the top-1 test accuracy is displayed in the title. The main difference is that the train loss \& train accuracy are in blue while the test loss \& test accuracy are in green. 
\subsubsection{Adam with no StepLR}
\begin{figure}[H]
    \centering
    \subfigure[Accuracy vs. Epochs]{\includegraphics[width=0.5\textwidth]{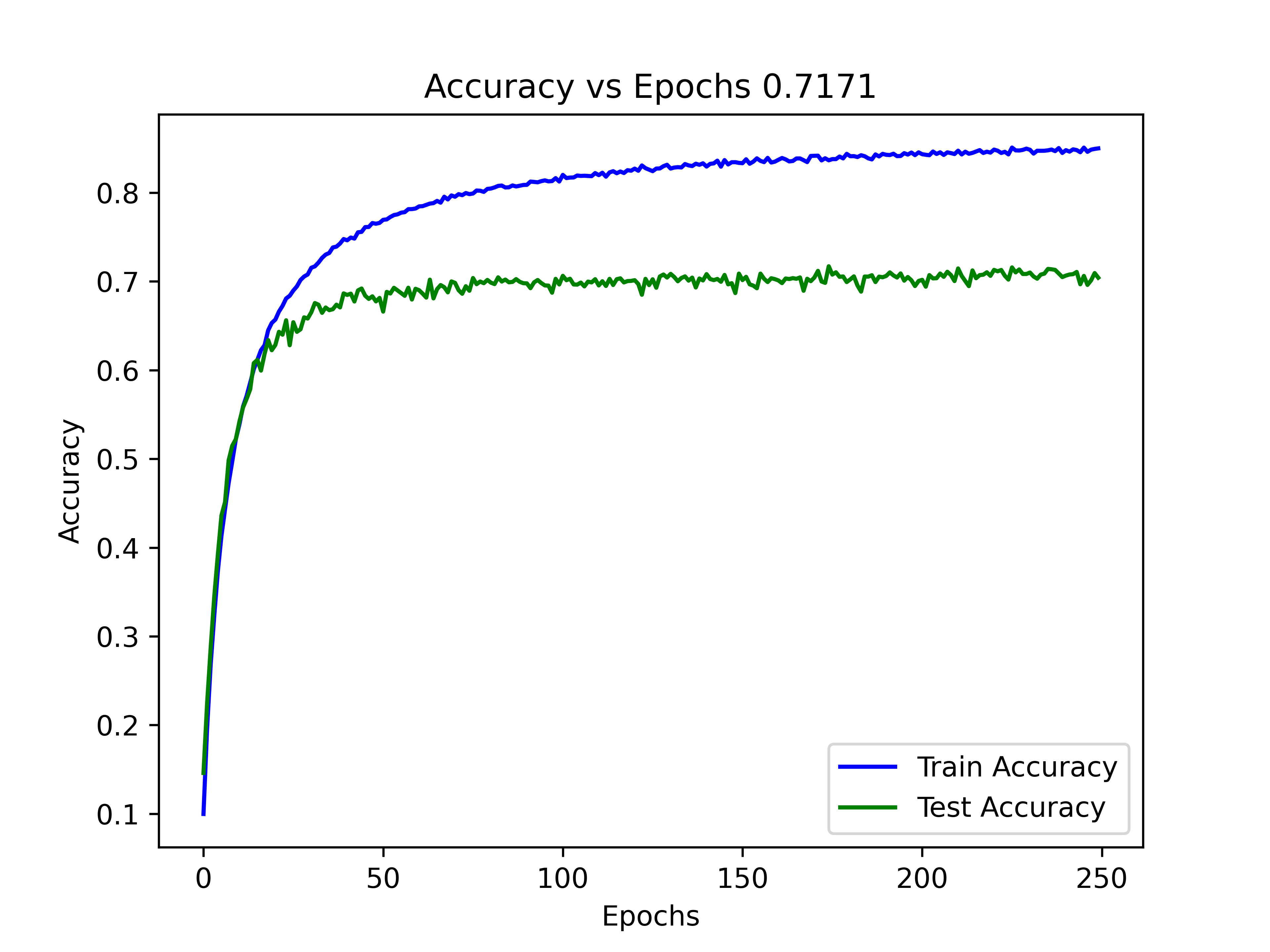}}\hspace{-0.80cm}
    \subfigure[Loss vs Epochs]{\includegraphics[width=0.5\textwidth]{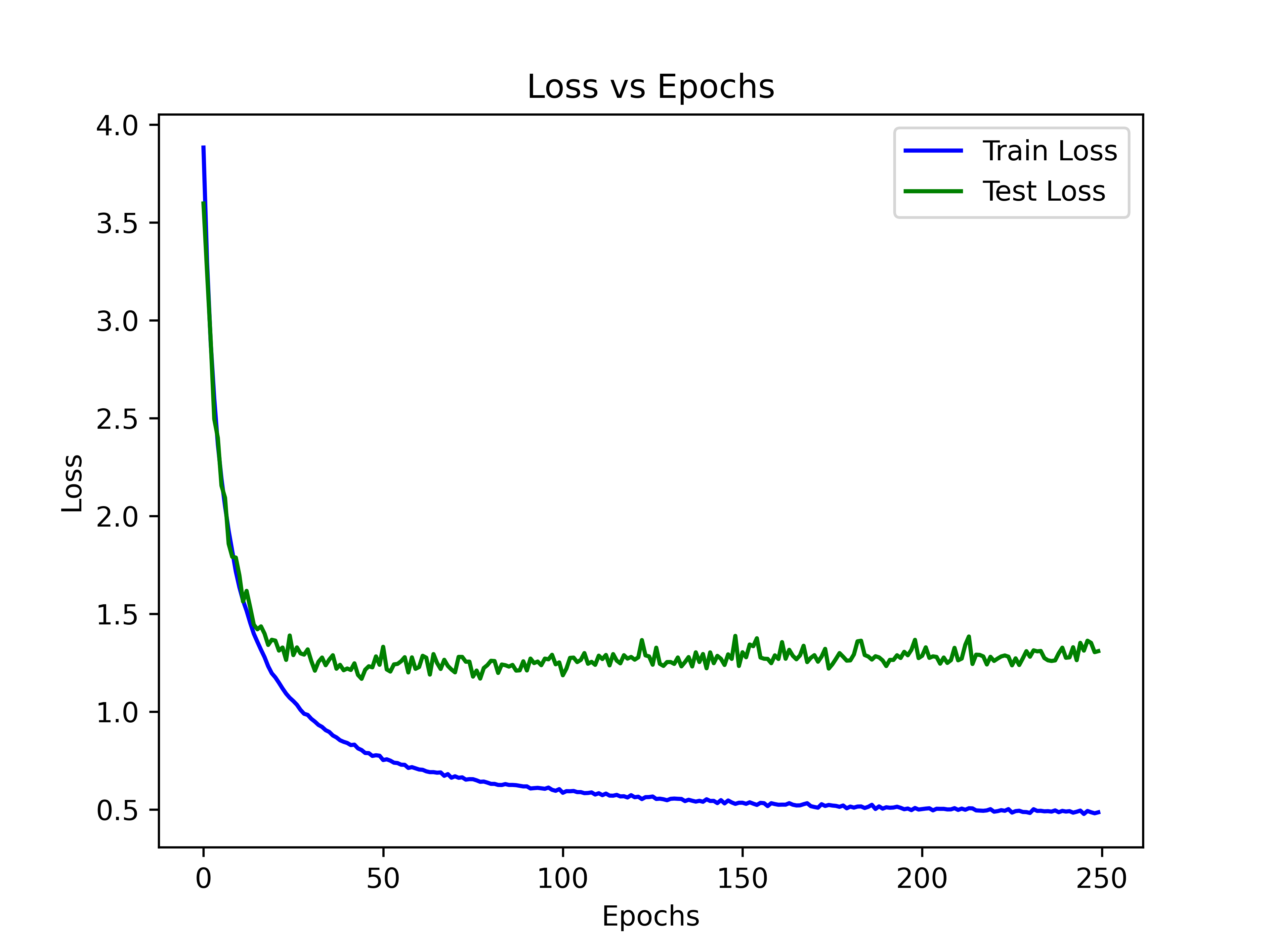}}
    \caption{Non-Federated ResNet with Adam and no StepLR}
    \label{fig:normal-adam-nosteplr}
\end{figure}

\subsubsection{Adam with StepLR}
\begin{figure}[H]
    \centering
    \subfigure[Accuracy vs. Epochs]{\includegraphics[width=0.5\textwidth]{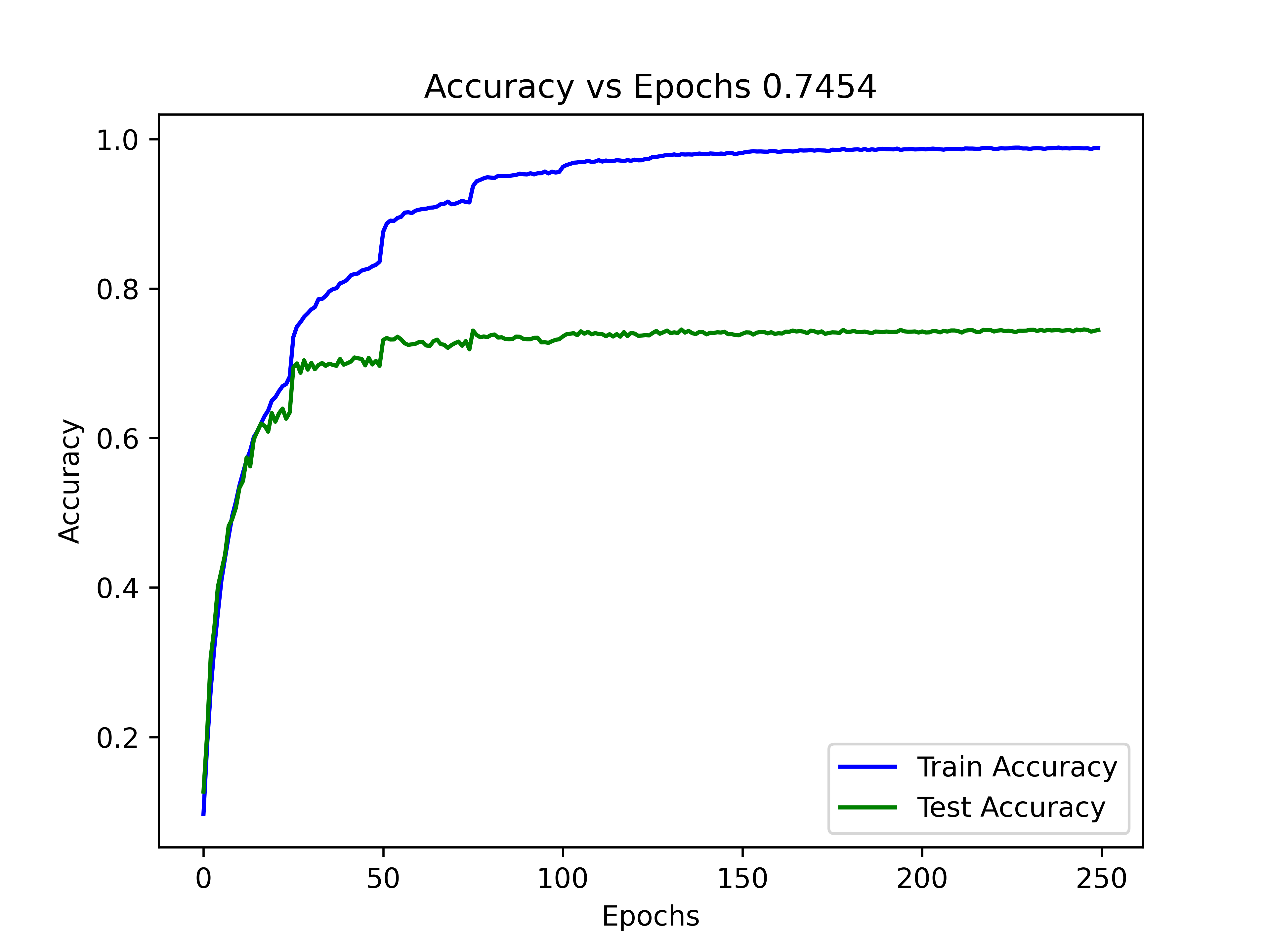}}\hspace{-0.80cm}
    \subfigure[Loss vs Epochs]{\includegraphics[width=0.5\textwidth]{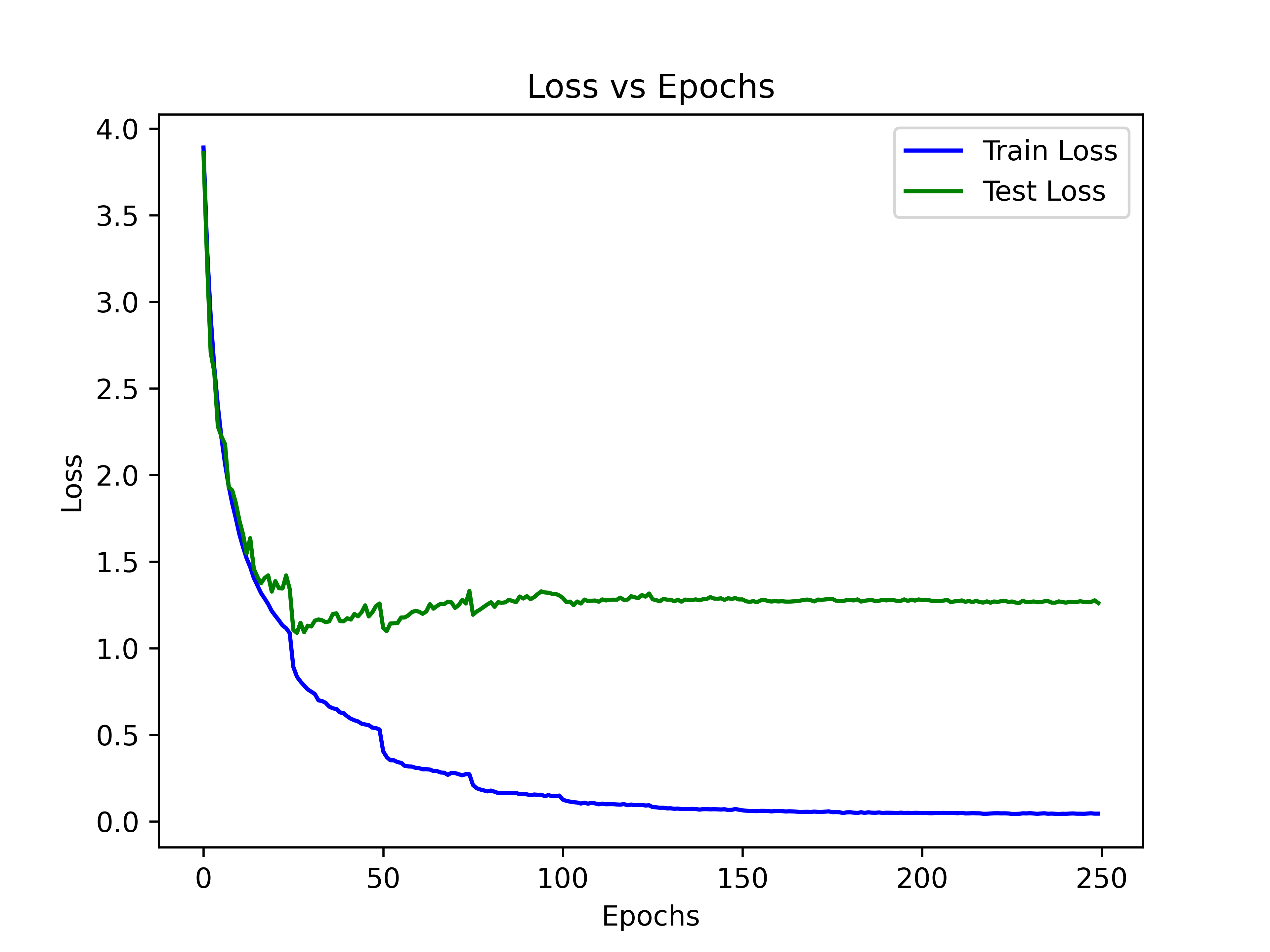}}
    \caption{Non-Federated ResNet18 with Adam and StepLR}
    \label{fig:normal-adam-steplr}
\end{figure}

\subsubsection{AdamP with no StepLR}
\begin{figure}[H]
    \centering
    \subfigure[Accuracy vs. Epochs]{\includegraphics[width=0.5\textwidth]{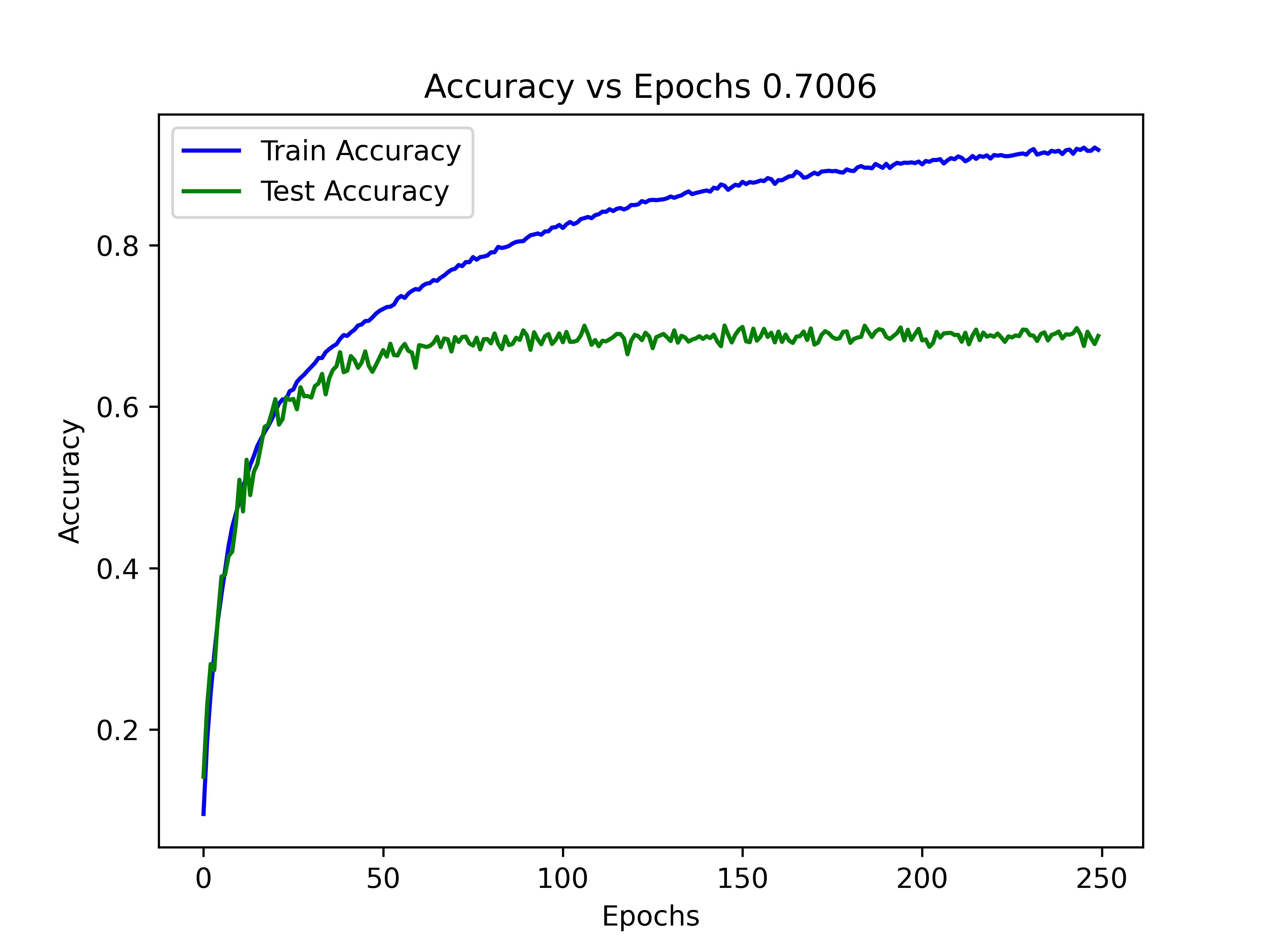}}\hspace{-0.80cm}
    \subfigure[Loss vs Epochs]{\includegraphics[width=0.5\textwidth]{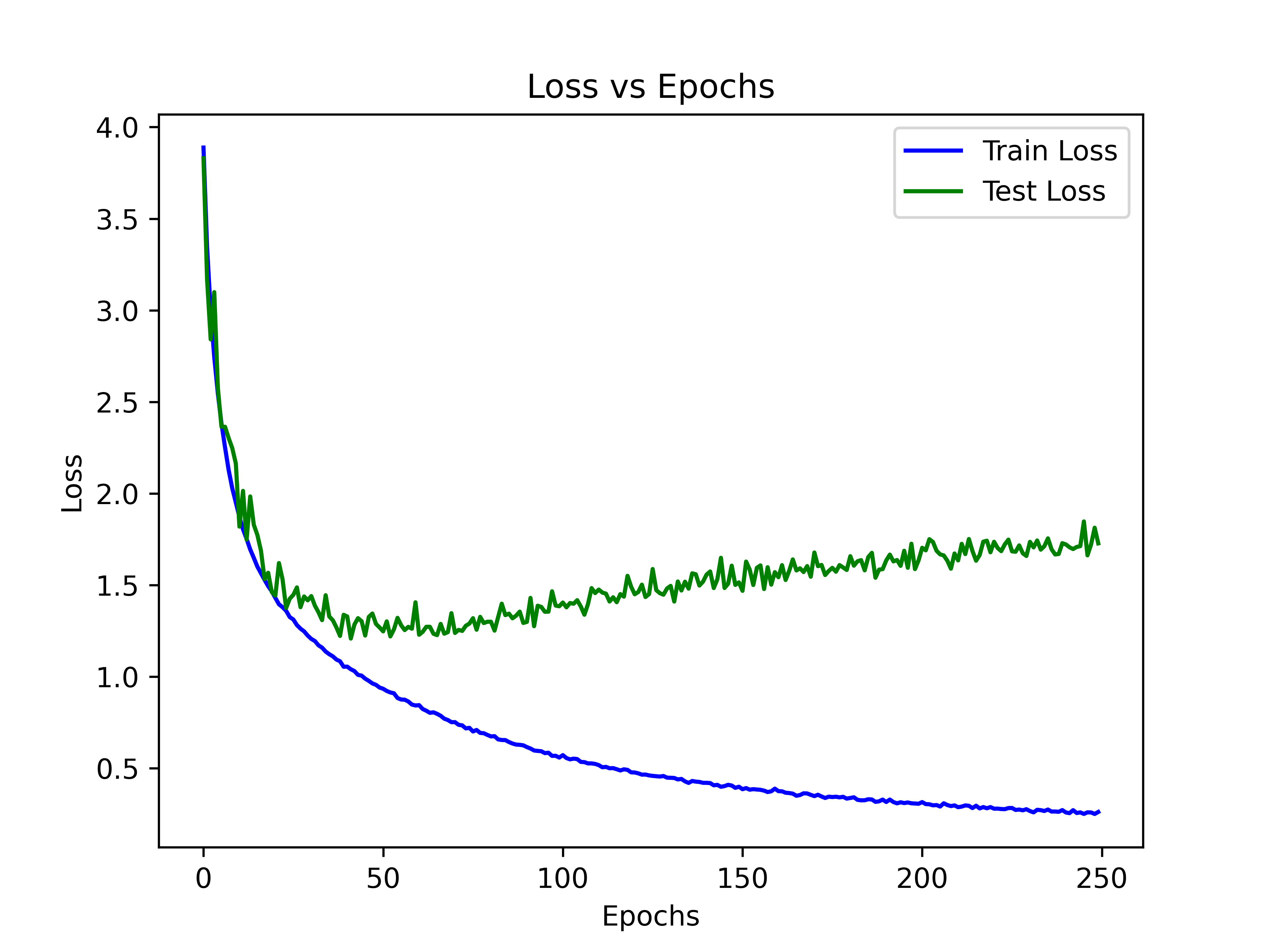}}
    \caption{Non-Federated ResNet18 with Adam and StepLR}
    \label{fig:normal-adamp-nosteplr}
\end{figure}

\subsubsection{AdamP with StepLR}
\begin{figure}[H]
    \centering
    \subfigure[Accuracy vs. Epochs]{\includegraphics[width=0.5\textwidth]{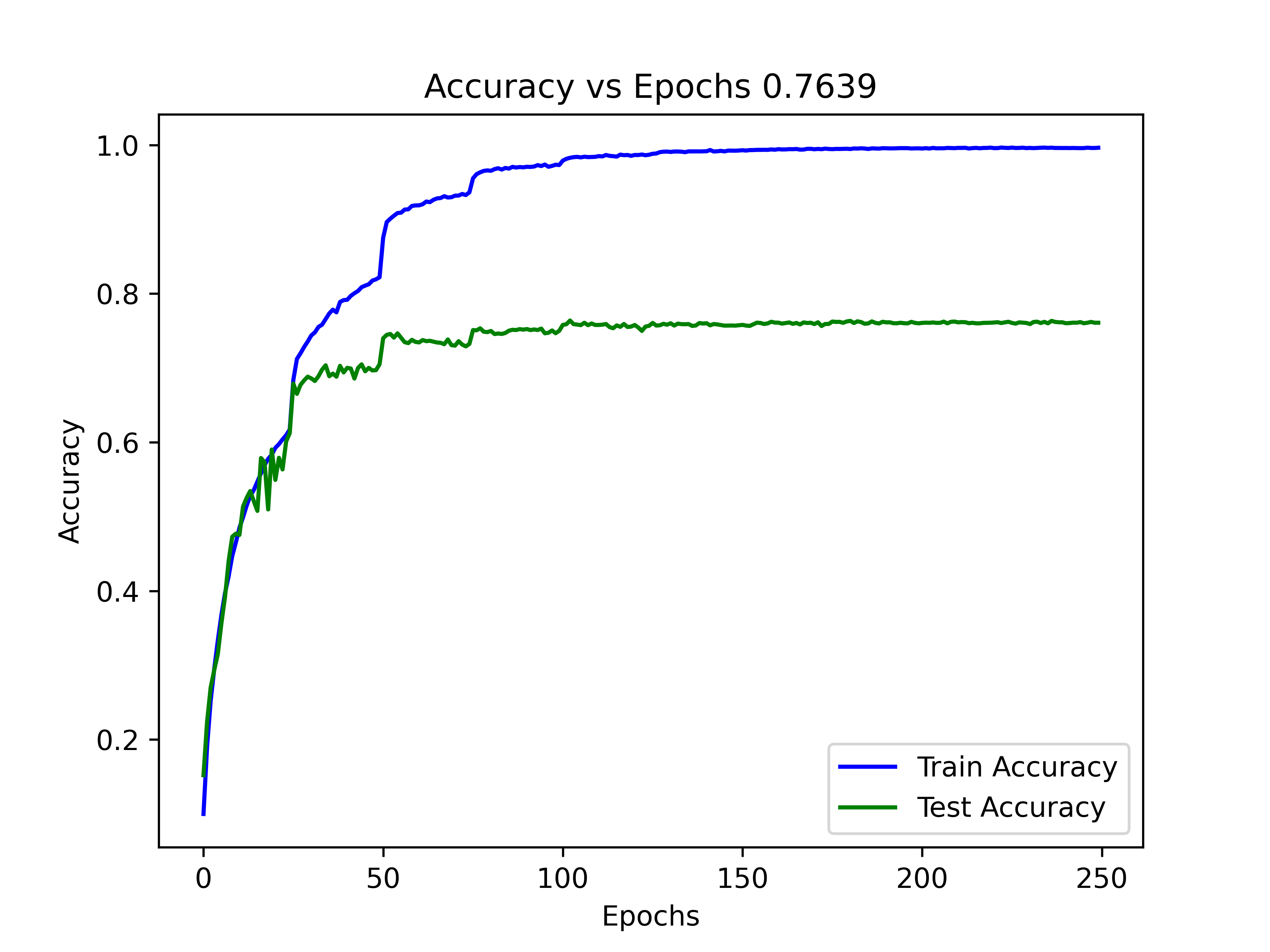}}\hspace{-0.80cm}
    \subfigure[Loss vs Epochs]{\includegraphics[width=0.5\textwidth]{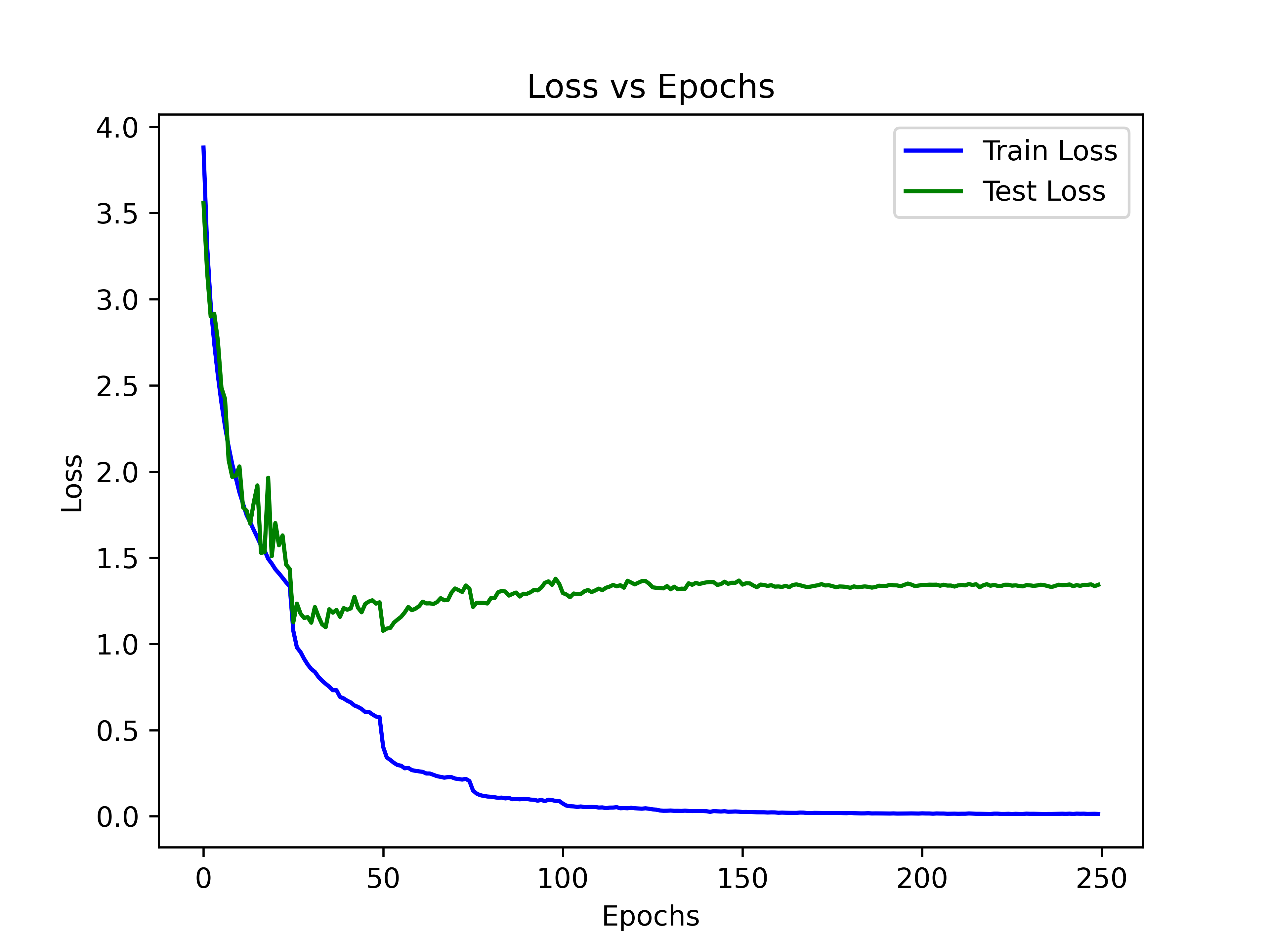}}
    \caption{Non-Federated ResNet18 with AdamP and StepLR}
    \label{fig:normal-adamp-steplr}
\end{figure}

\subsubsection{RMSGD with no StepLR}
\begin{figure}[H]
    \centering
    \subfigure[Accuracy vs. Steps]{\includegraphics[width=0.5\textwidth]{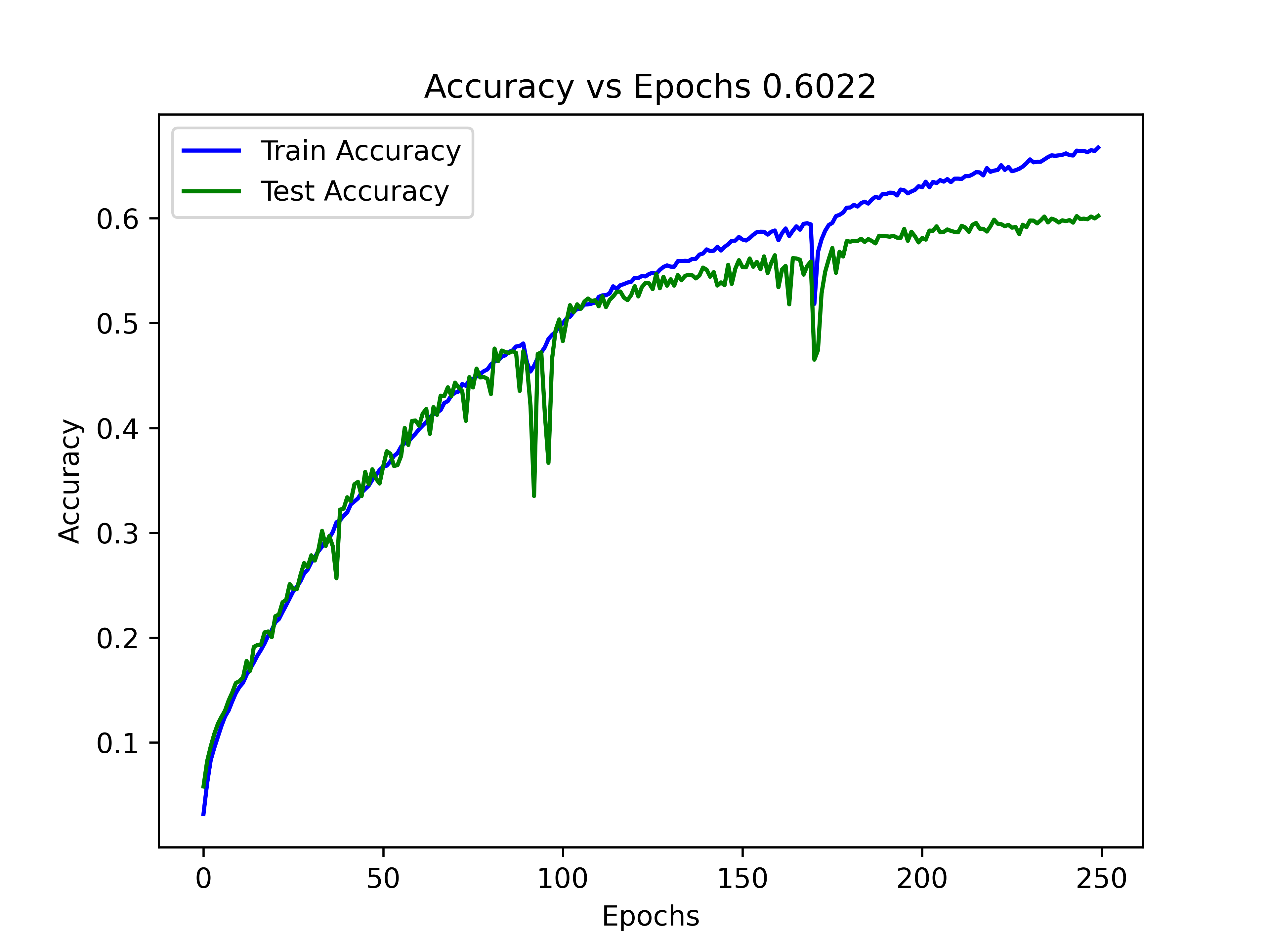}}\hspace{-0.80cm}
    \subfigure[Loss vs Steps]{\includegraphics[width=0.5\textwidth]{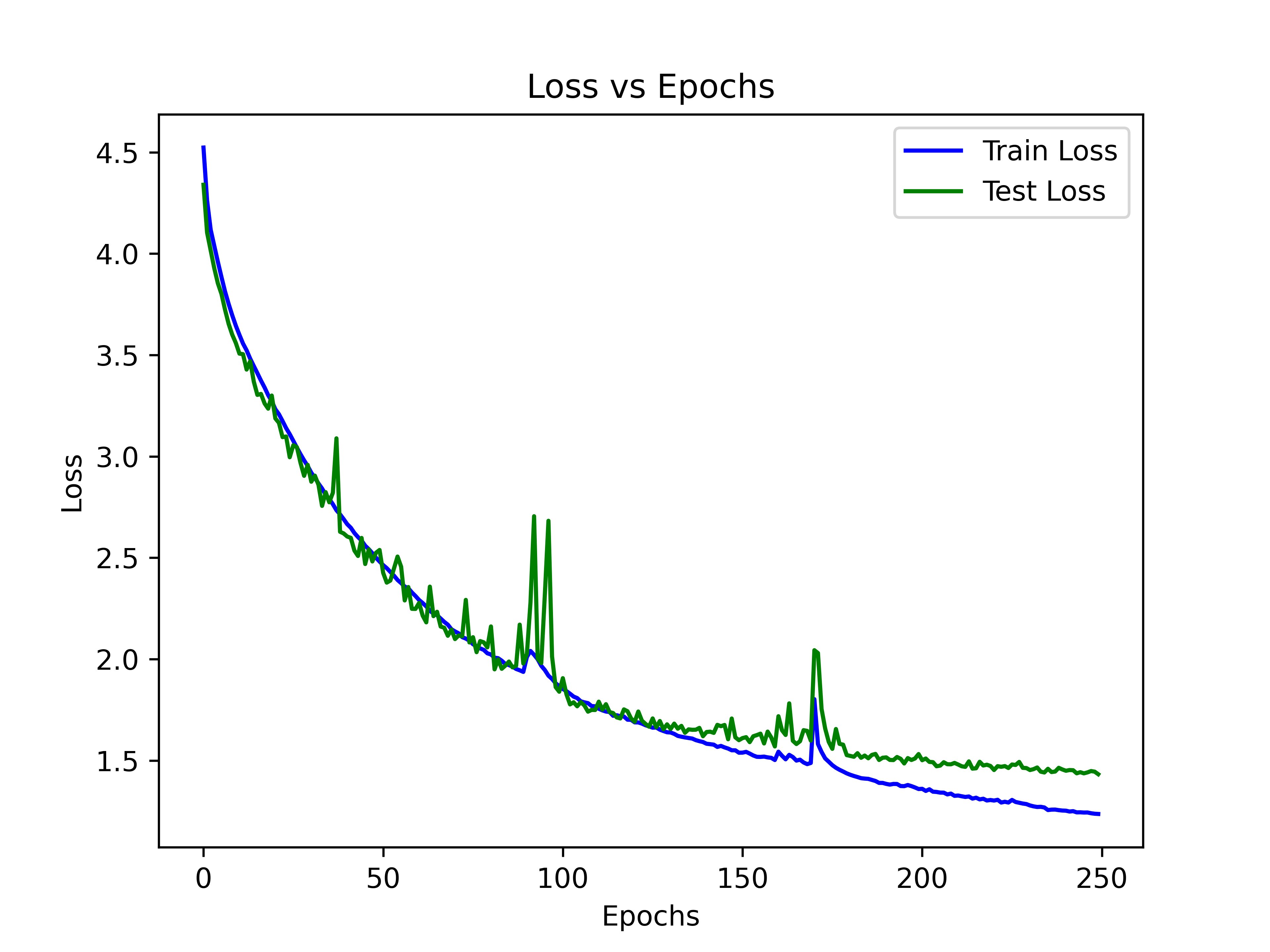}}
    \caption{Non-Federated ResNet18 with RMSGD and no StepLR}
    \label{fig:normal-rmsgd-nosteplr}
\end{figure}

\newpage
\subsection{Federated Averaging \cite{konevcny2016federated}}
\subsubsection{Adam with no StepLR}

\begin{figure}[H]
    \centering
    \subfigure[Accuracy vs. Steps]{\includegraphics[width=0.5\textwidth]{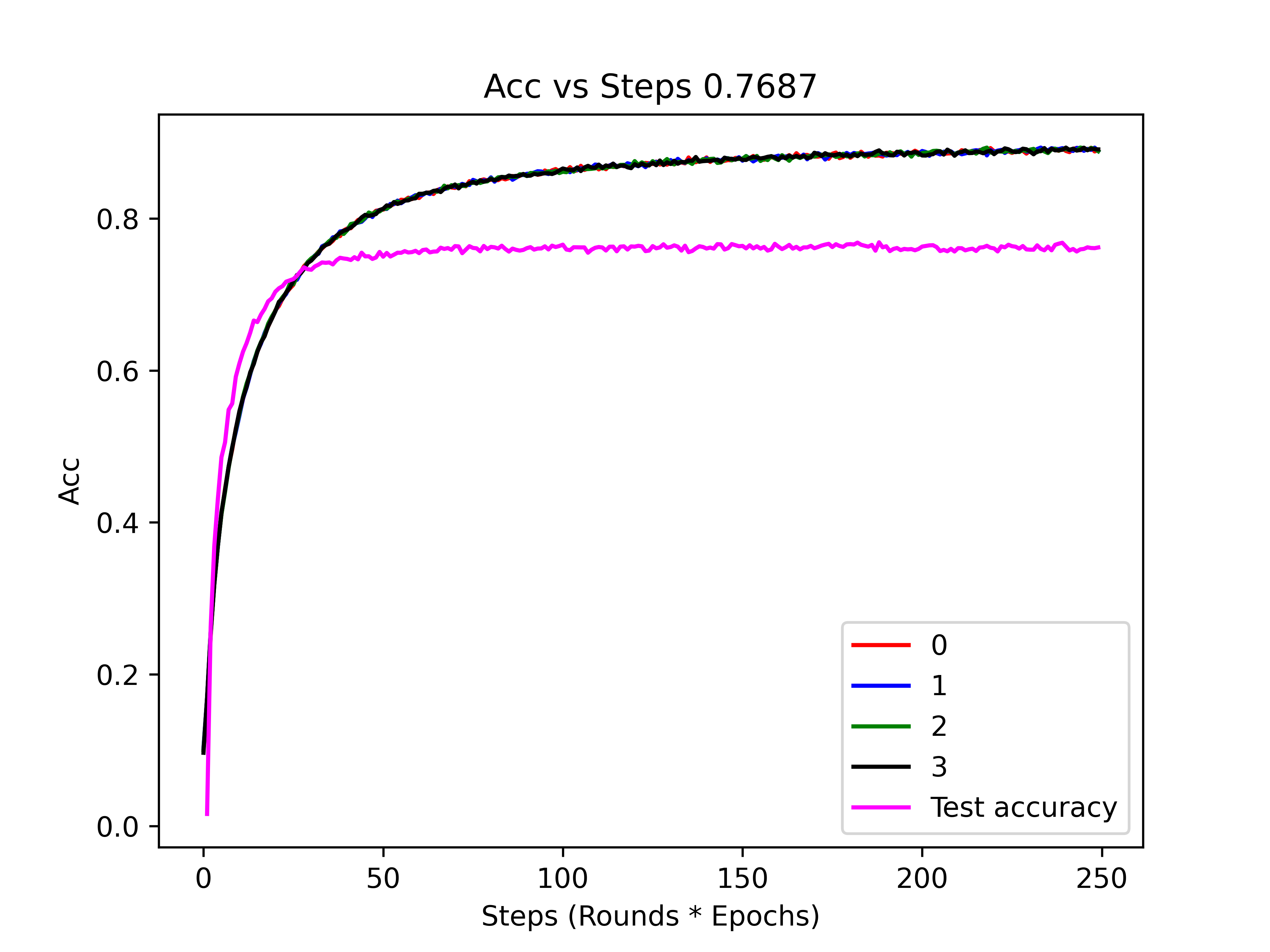}}\hspace{-0.80cm}
    \subfigure[Loss vs Steps]{\includegraphics[width=0.5\textwidth]{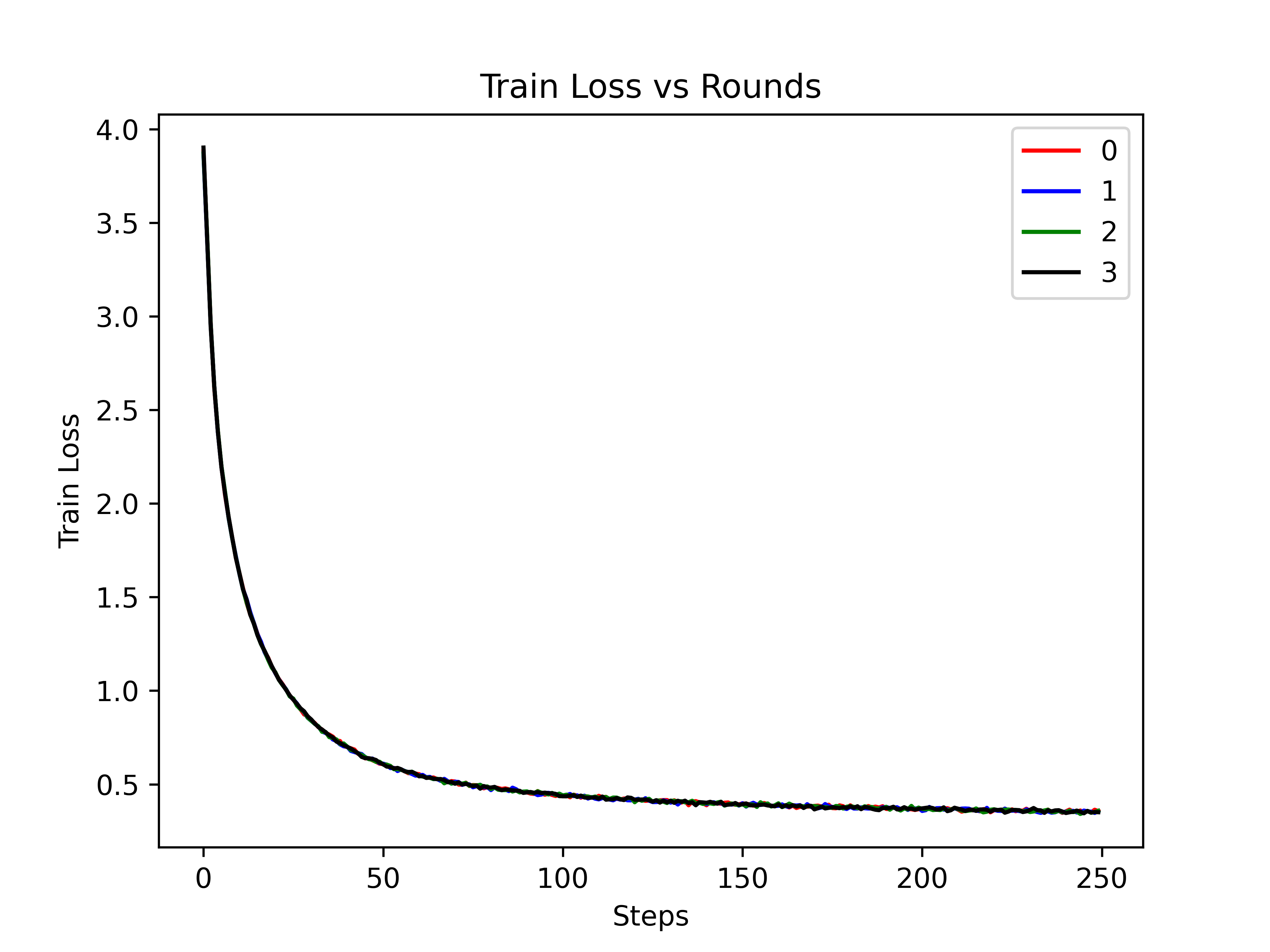}}
    \caption{Federated Averaging with Adam and no StepLR}
    \label{fig:fedavg-adam-nosteplr}
\end{figure}

Here, We can see the traditional accuracy \& loss curves. Due to the knowledge sharing nature we are leveraging in Federated Learning, Adam without StepLR is out-performing plain ResNet18 on CIFAR100 seen in \autoref{fig:normal-adam-nosteplr}.
 
\newpage
\subsubsection{Adam with StepLR}

\begin{figure}[H]
    \centering
    \subfigure[Accuracy vs. Steps]{\includegraphics[width=0.5\textwidth]{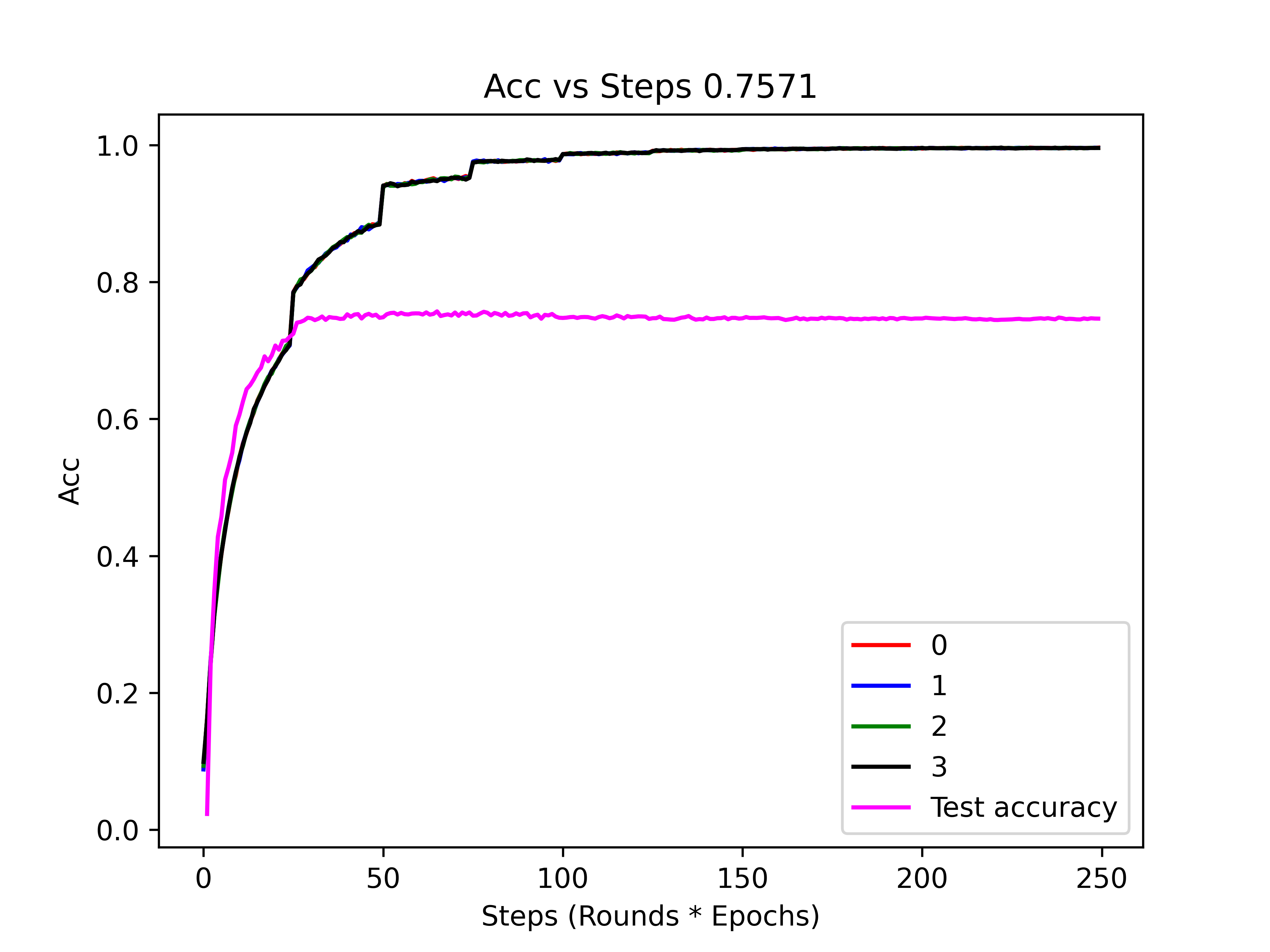}}\hspace{-0.80cm}
    \subfigure[Loss vs Steps]{\includegraphics[width=0.5\textwidth]{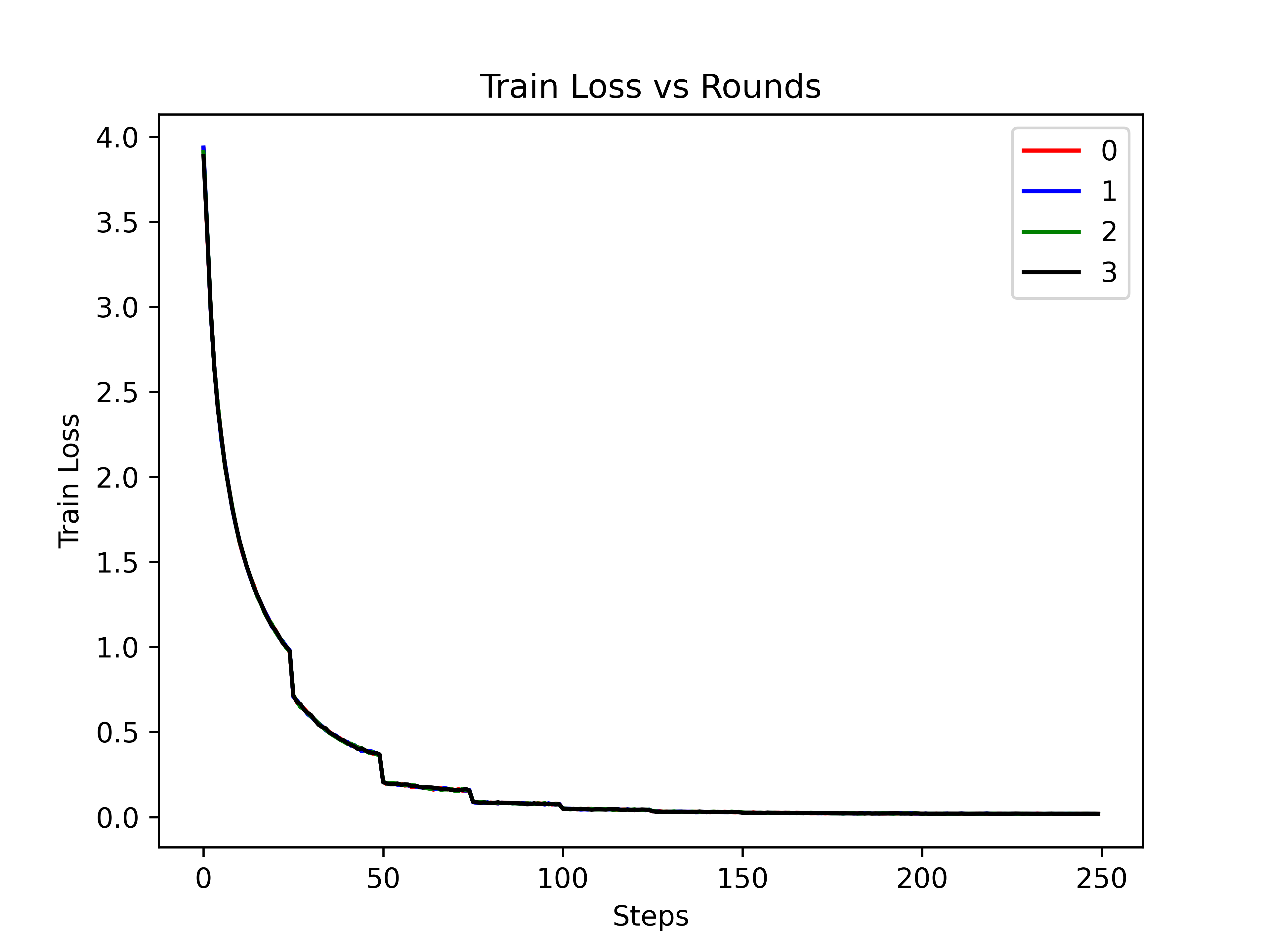}}
    \caption{Federated Averaging with Adam and StepLR}
    \label{fig:fedavg-adam-steplr}
\end{figure}

In comparison to \autoref{fig:fedavg-adam-nosteplr}, we can see a visible increase in the generalization gap with an under-performing test-accuracy. This is contradictory to conventional wisdom, where lowering the learning rate should prevent oscillations in the loss landscape and thus encourage learning, but here as the training accuracy increases there is not a proportional increase in the test accuracy, suggesting overfitting. 

\newpage
\subsubsection{AdamP with no StepLR}

\begin{figure}[H]
    \centering
    \subfigure[Accuracy vs. Steps]{\includegraphics[width=0.5\textwidth]{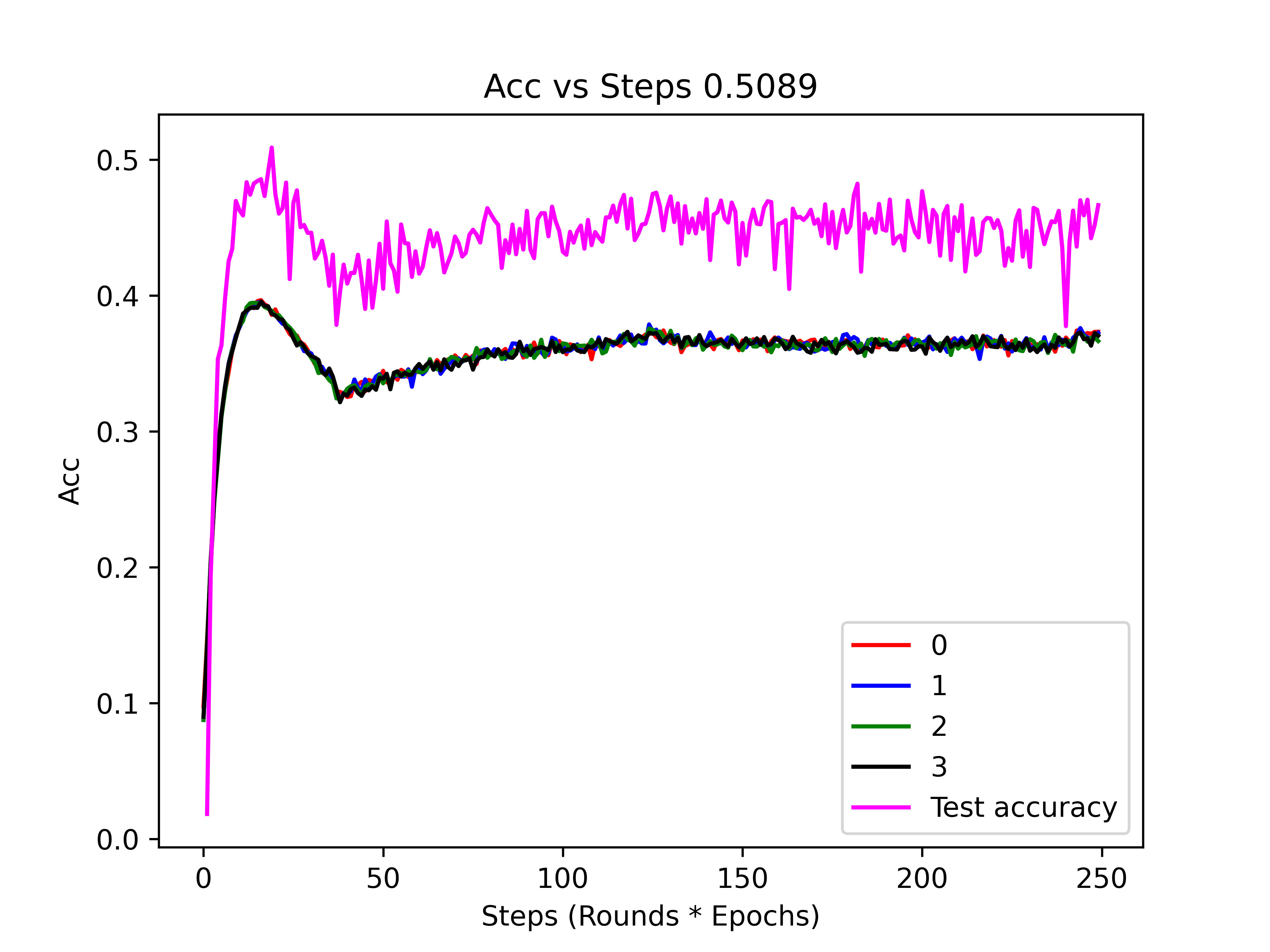}}\hspace{-0.80cm}
    \subfigure[Loss vs Steps]{\includegraphics[width=0.5\textwidth]{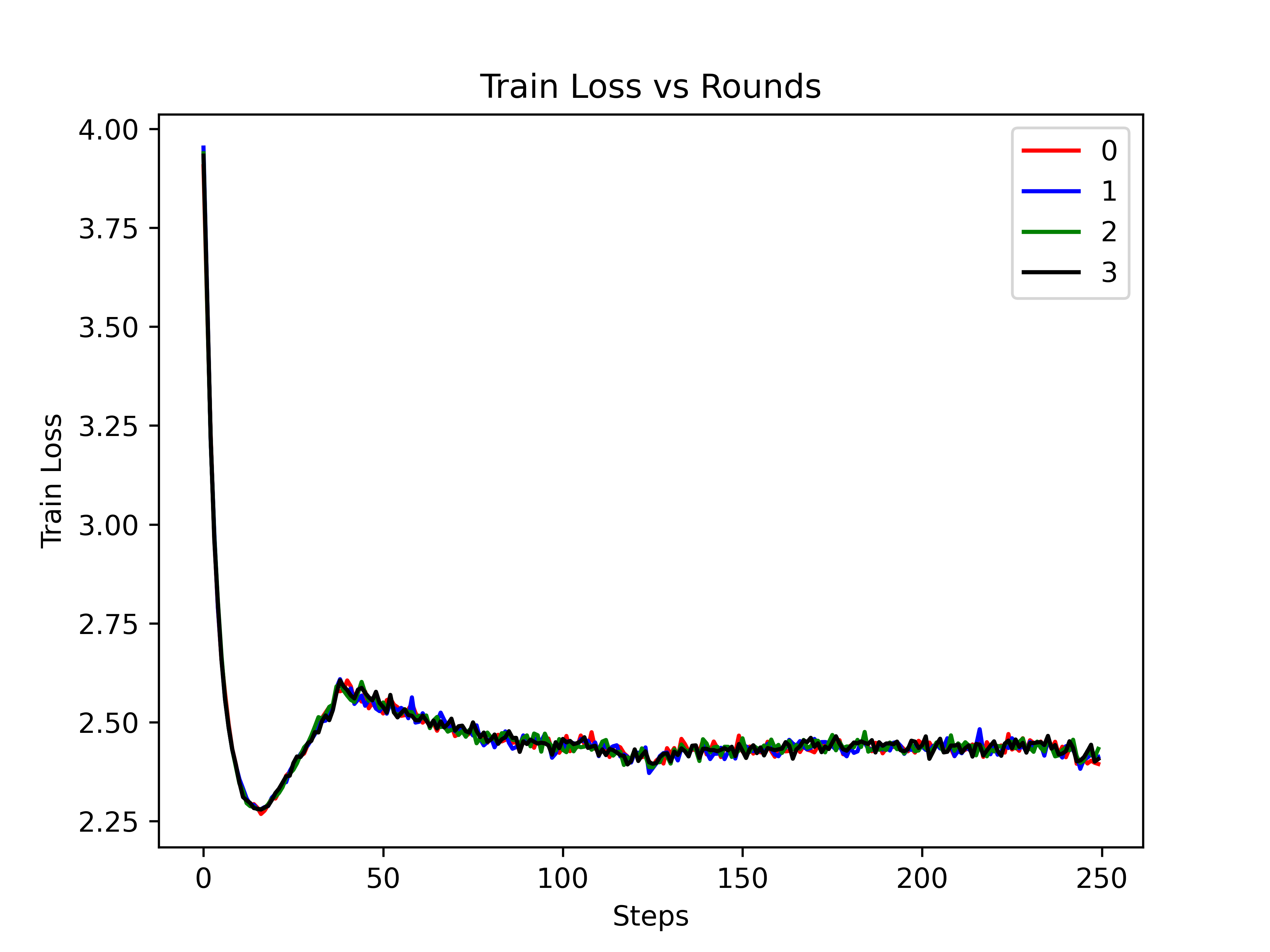}}
    \caption{Federated Averaging with AdamP and no StepLR}
    \label{fig:fedavg-adamp-nosteplr}
\end{figure}

The figure in \autoref{fig:fedavg-adamp-nosteplr} demonstrates a unique phenomena, an incredibly rapid growth within the first regime (first 15 epochs) then a come-down in accuracy (identical increase in loss) then a struggle to lower loss again. This could be due to a bad initial seed, but since all four clients start on different seeds to aggregate knowledge, this seems unlikely.\\

As mentioned in \autoref{sec:optimizers} AdamP is a projection onto the tangent of the gradient. I suspect that when aggregating the results, that AdamP's is \textit{over-regularizing} the weights. This is further supported by the train accuracy being lower than the test-accuracy. \\

Therefore, further research into this phenomena should be considered, changing \textbf{a)} the learning rate to something lower \& perhaps using StepLR (addressed in the next section)  \textbf{b)} lowering the weight-decay parameter (although the weight-decay chosen is actually lower than the default) \textbf{c)} increase the $\beta_1$ \& $\beta_2$ parameters in Adam to encourage averaging over a larger history, allowing for increased smoothness. 

\subsubsection{AdamP with StepLR}

\begin{figure}[H]
    \centering
    \subfigure[Accuracy vs. Steps]{\includegraphics[width=0.5\textwidth]{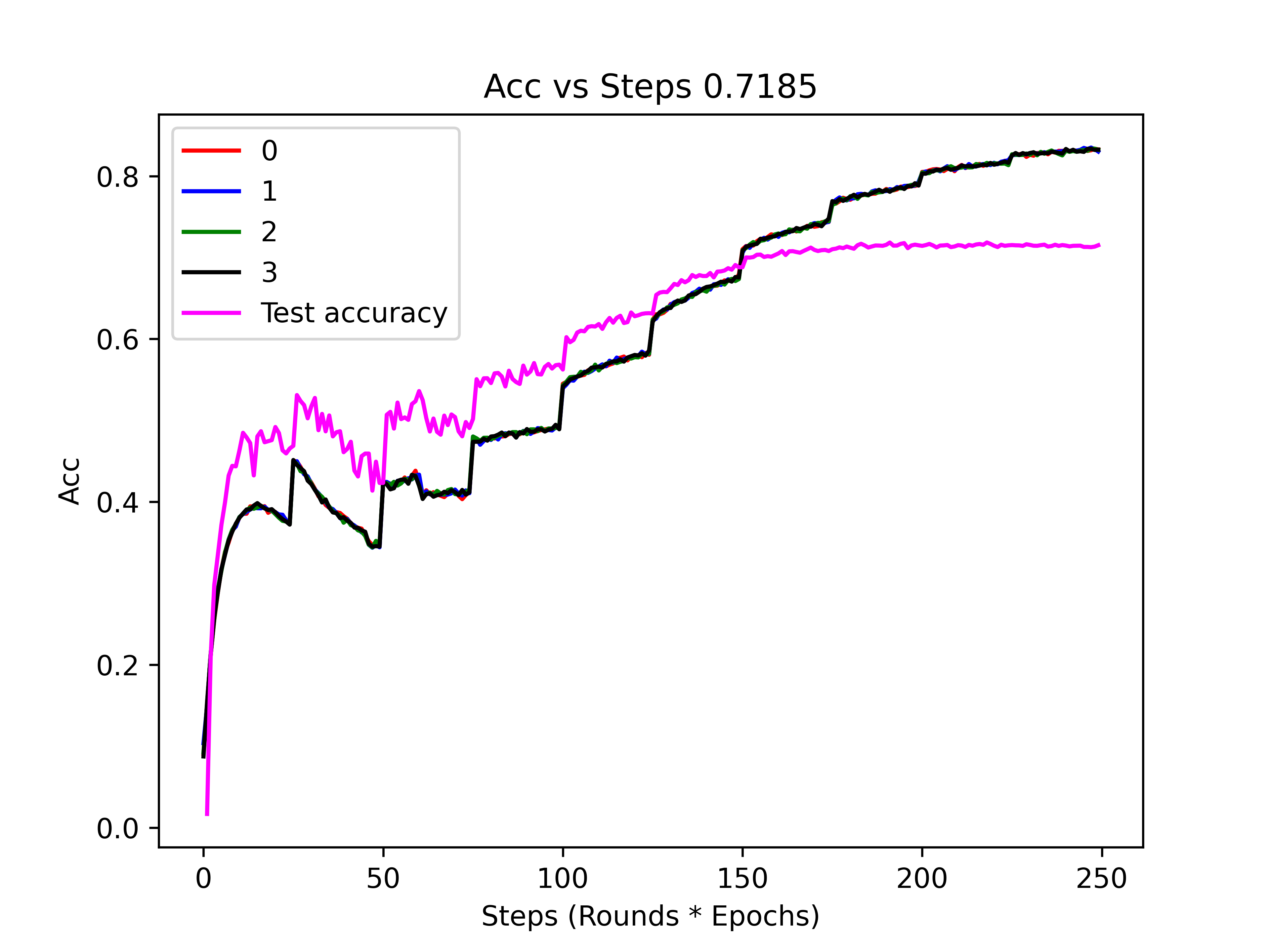}}\hspace{-0.80cm}
    \subfigure[Loss vs Steps]{\includegraphics[width=0.5\textwidth]{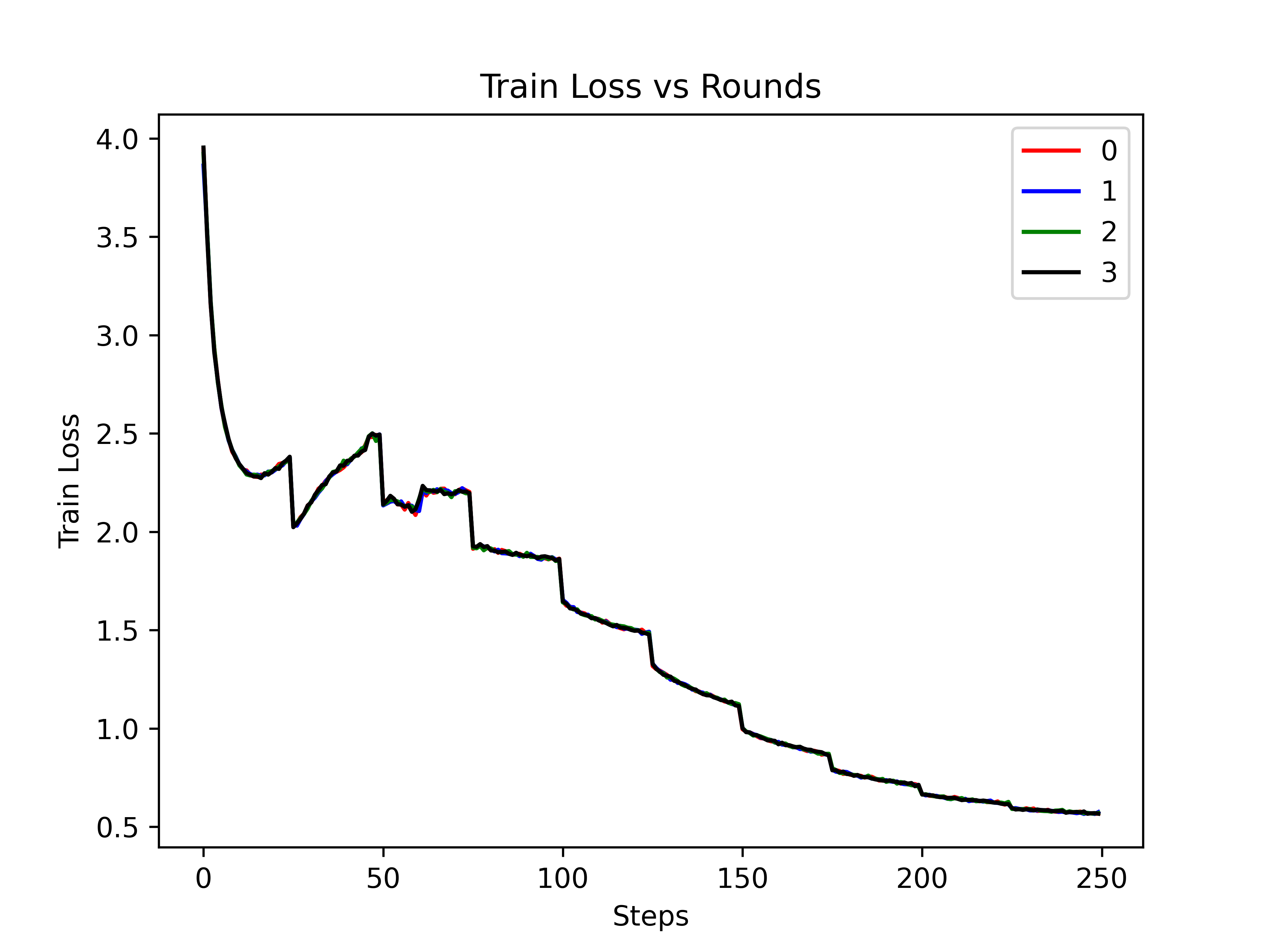}}
    \caption{Federated Averaging with AdamP and StepLR}
    \label{fig:fedavg-adamp-steplr}
\end{figure}

Using StepLR has a significant change to AdamP's usage in a federated setting. When comparing to \autoref{fig:fedavg-adamp-nosteplr} we can observe how it simultaneously suffers from the same over-regularization, as the test-accuracy remains higher than the train-accuracy, but the StepLR forcing the train-accuracy up (\& loss down) to overcome the regularization in order to get to a competitive test-accuracy. It appears that the regularization \& overfitting of StepLR are `fighting' to determine which one dominates the model. Furthermore, the fast initial learning suggests value can be extracted from this method.
\newpage
\subsubsection{RMSGD with no StepLR}

\begin{figure}[H]
    \centering
    \subfigure[Accuracy vs. Steps]{\includegraphics[width=0.5\textwidth]{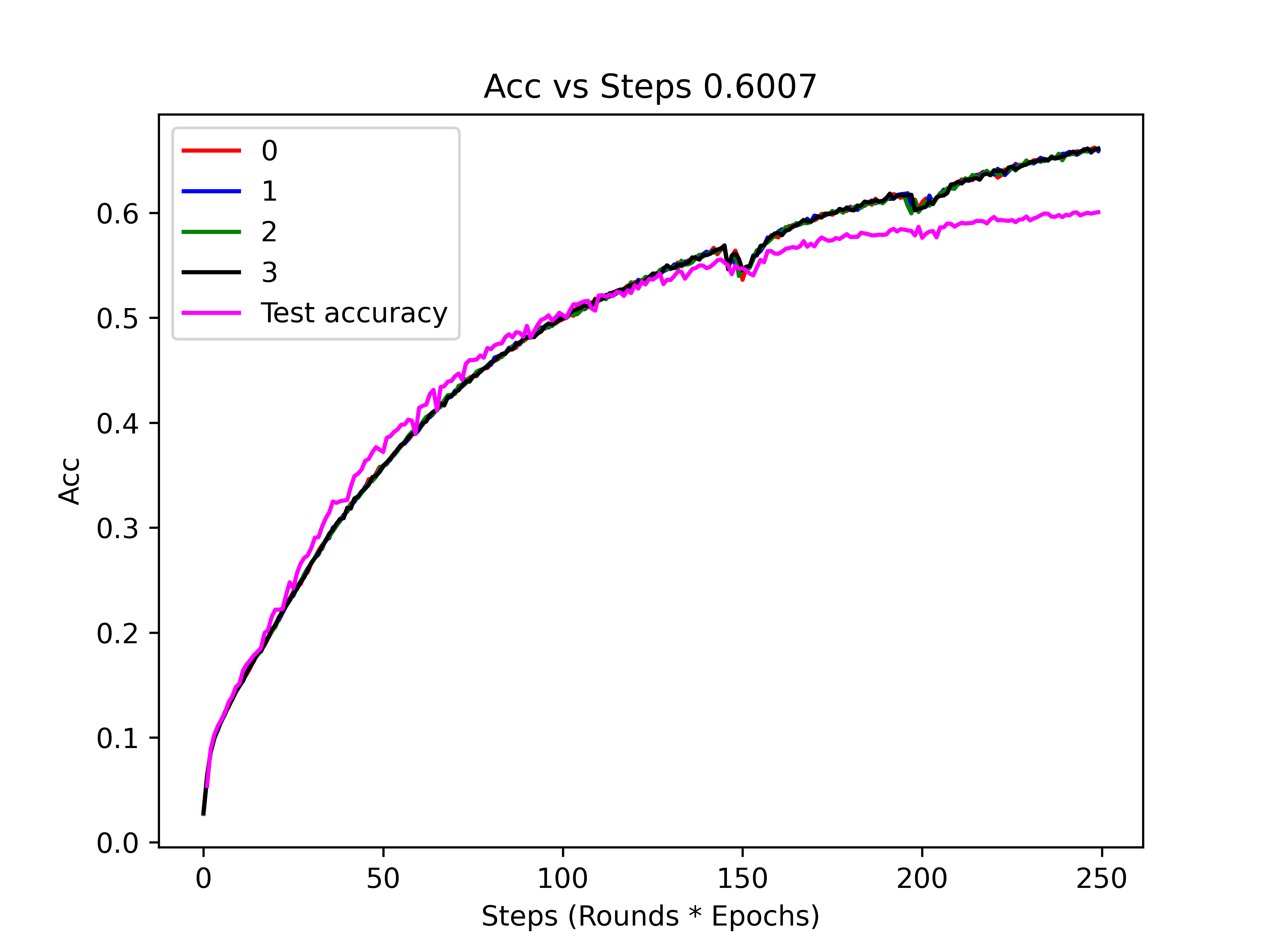}}\hspace{-0.80cm}
    \subfigure[Loss vs Steps]{\includegraphics[width=0.5\textwidth]{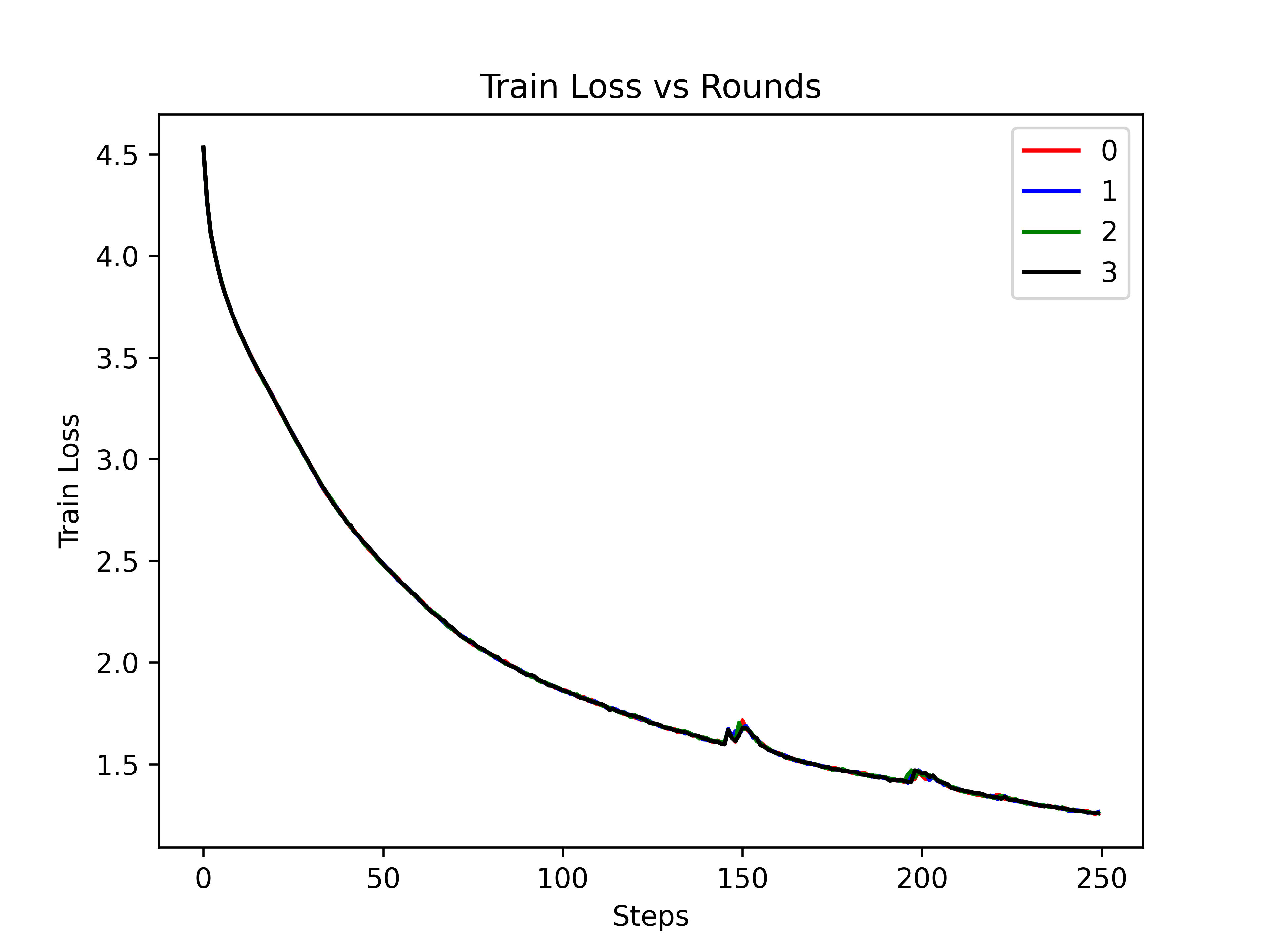}}
    \caption{Federated Averaging with RMSGD and no StepLR}
    \label{fig:fedavg-rmsgd-nosteplr}
\end{figure}

While RMSGD does under-perform the other models, it simultaneously has the lowest generalization-gap across all possible variations. The most fascinating thing about \autoref{fig:fedavg-rmsgd-nosteplr} is the rate at which it learns is extremely slowly. Since it is learning at such a slow pace, it can be ascertained that the model's learning rate was set too low (1e-3) and could benefit from \textbf{a)} longer training times and \textbf{b)} a higher learning rate, possibly with a learning rate scheduler like StepLR. With these key changes, further exploration into how RMSGD could improve in the federated setting should be considered.

\newpage
\subsection{Federated Effective Rank}
\subsubsection{Adam with no StepLR}
\begin{figure}[H]
    \centering
    \subfigure[Accuracy vs. Steps]{\includegraphics[width=0.5\textwidth]{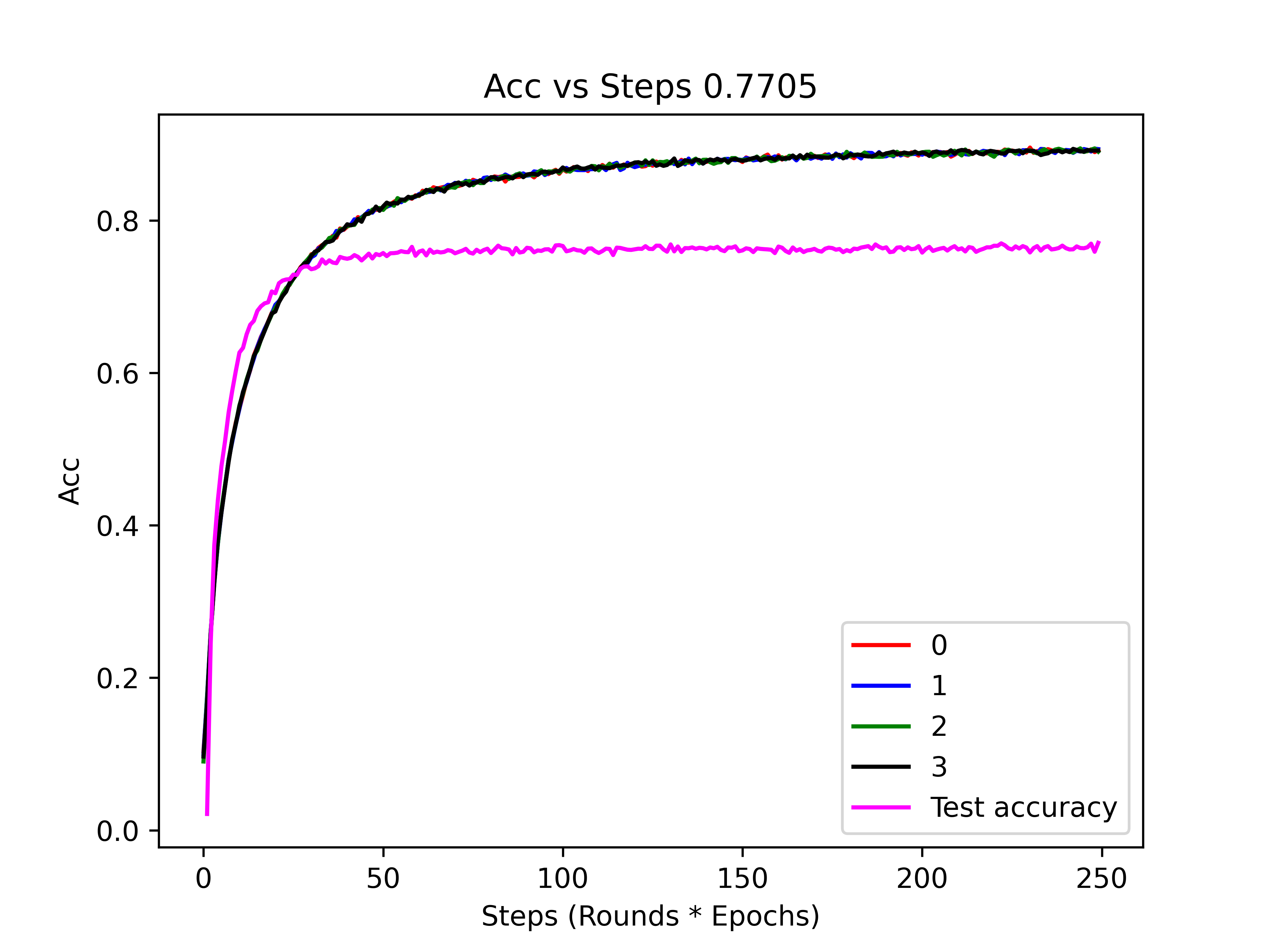}}\hspace{-0.80cm}
    \subfigure[Loss vs Steps]{\includegraphics[width=0.5\textwidth]{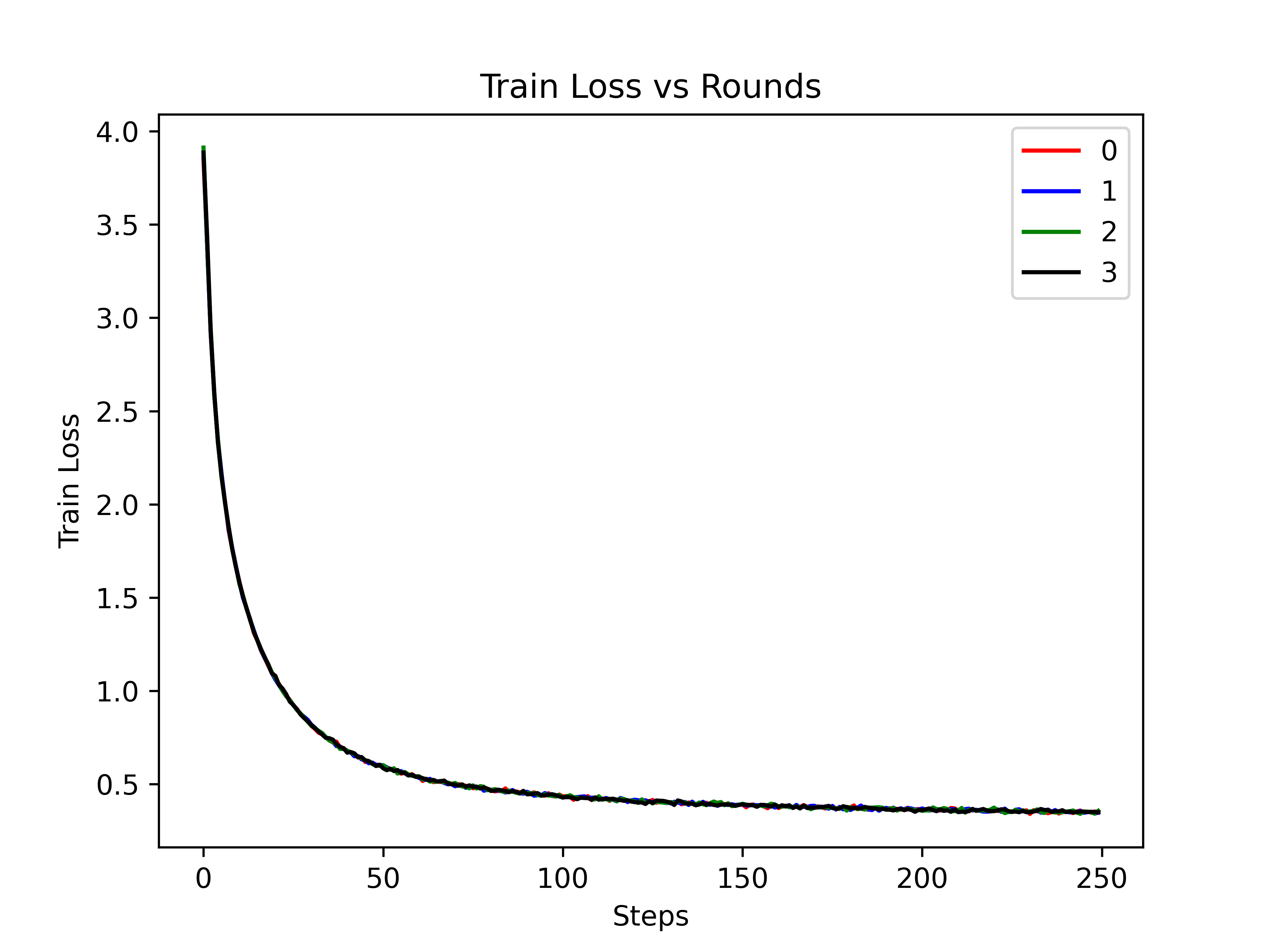}}
    \caption{Federated Effective Rank (FedER) with Adam and no StepLR}
    \label{fig:feder-adam-nosteplr}
\end{figure}

The Federated Effective Rank (FedER) method out-performs \autoref{fig:fedavg-adam-nosteplr} by \textcolor{ao}{\textbf{+0.18\%}}. As mentioned in \autoref{sec:er}, the effective rank is a form of measuring the feature mapping representations, and by utilizing the effective rank to weight-convolutional weight aggregations, does indeed prove useful in a federated setting.

\newpage
\subsubsection{Adam with StepLR}

\begin{figure}[H]
    \centering
    \subfigure[Accuracy vs. Steps]{\includegraphics[width=0.5\textwidth]{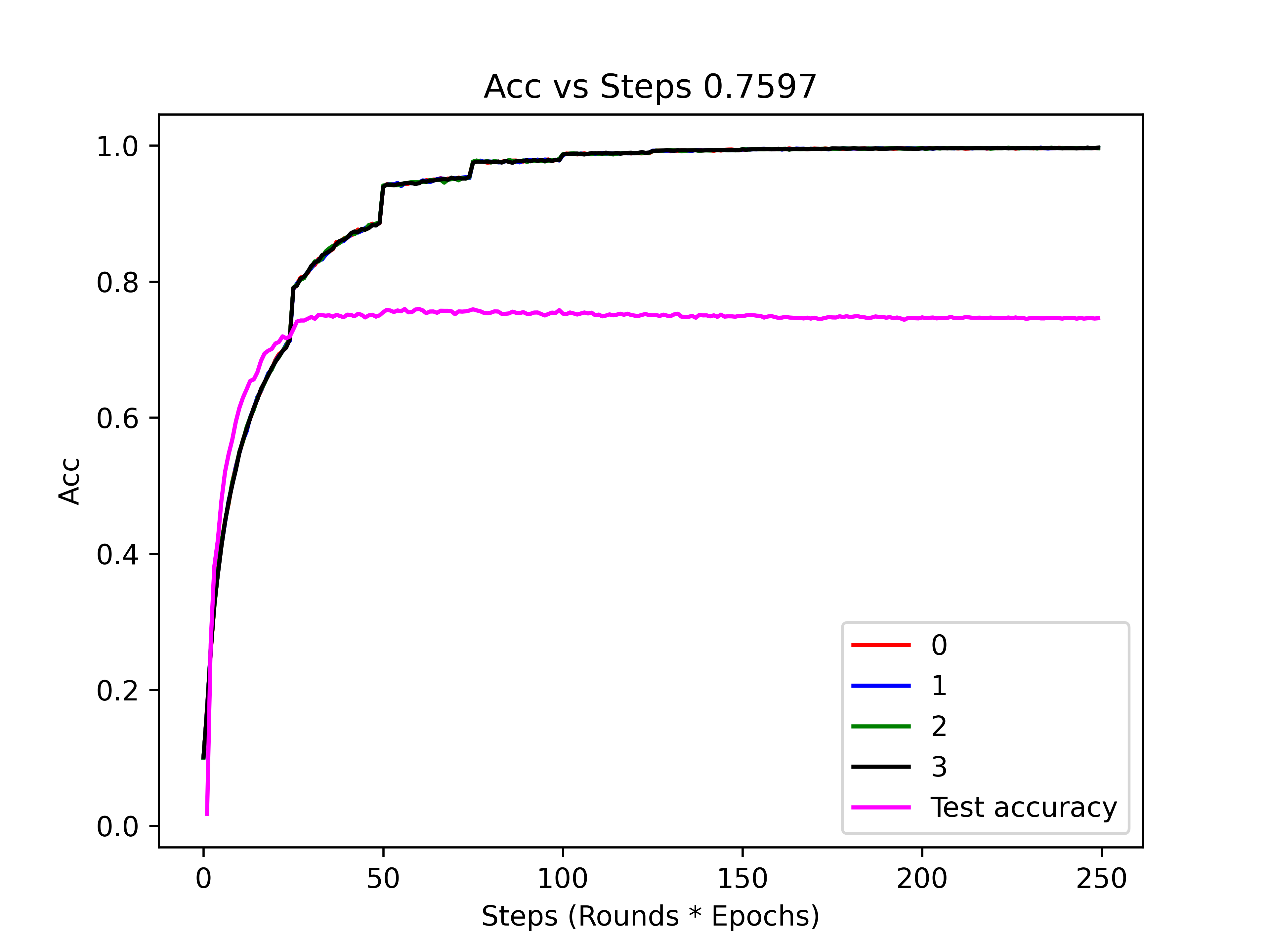}}\hspace{-0.80cm}
    \subfigure[Loss vs Steps]{\includegraphics[width=0.5\textwidth]{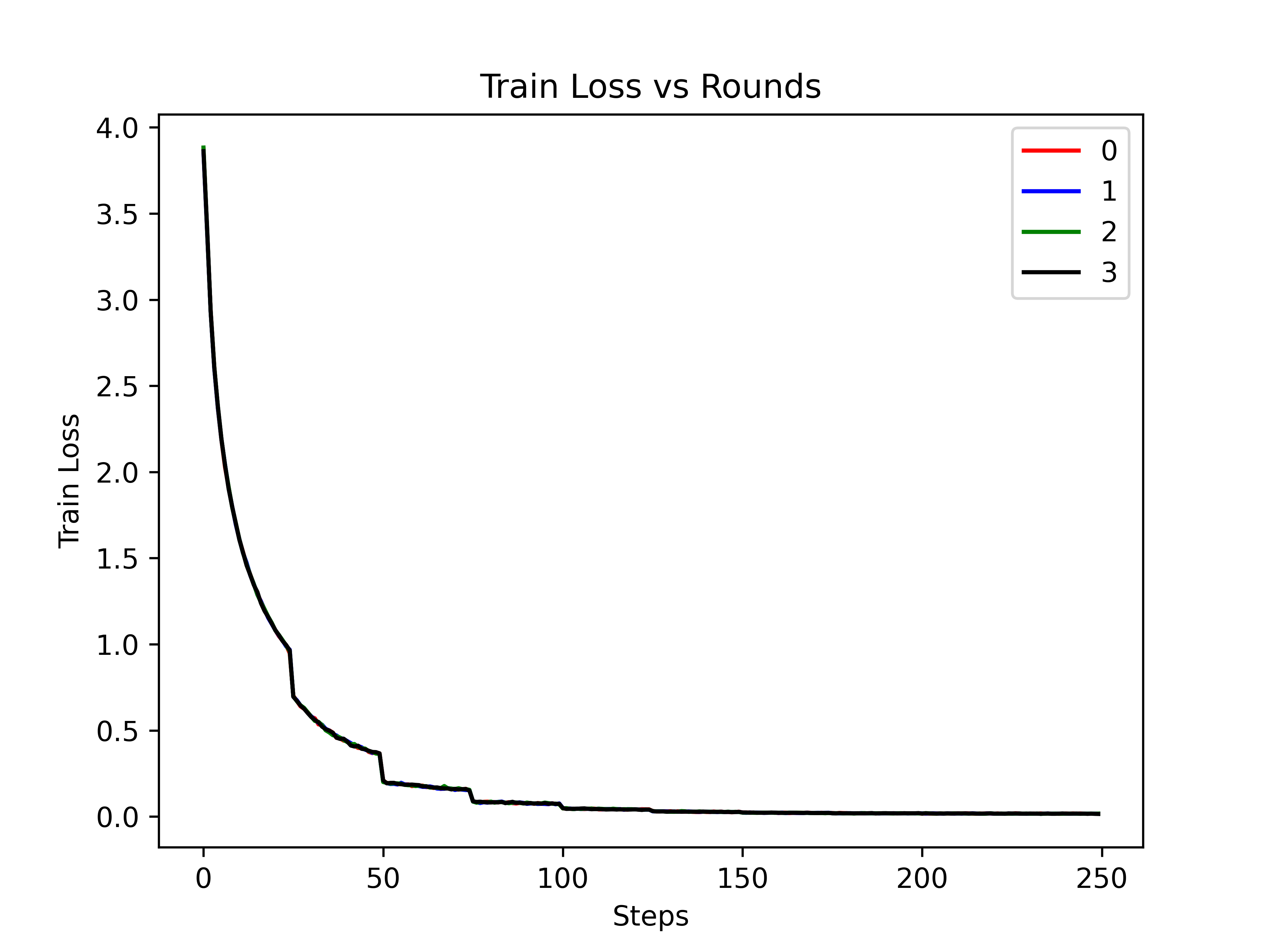}}
    \caption{Federated Effective Rank (FedER) with Adam and StepLR}
    \label{fig:feder-adam-steplr}
\end{figure}

When using StepLR, the same issue of over-fitting plagues the FedER as did in FedAvg in \autoref{fig:fedavg-adam-steplr} such as the large generalization gap. That being said, this novel method, FedER outperforms the FedAvg by \textcolor{ao}{+0.26\%}, demonstrating how effective rank's mapping technique is state-of-the-art. 
\newpage
\subsubsection{AdamP with no StepLR}

\begin{figure}[H]
    \centering
    \subfigure[Accuracy vs. Steps]{\includegraphics[width=0.5\textwidth]{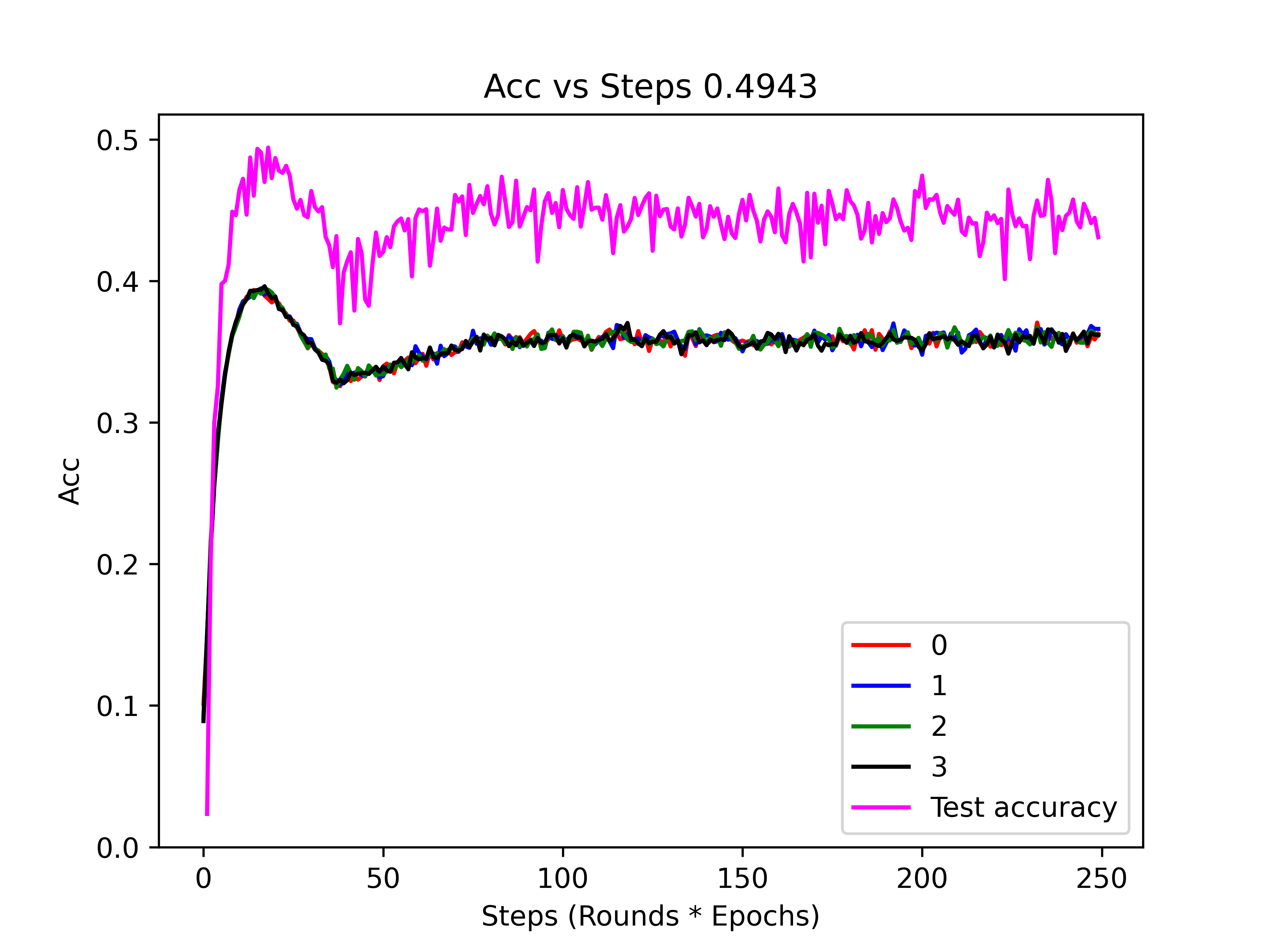}}\hspace{-0.80cm}
    \subfigure[Loss vs Steps]{\includegraphics[width=0.5\textwidth]{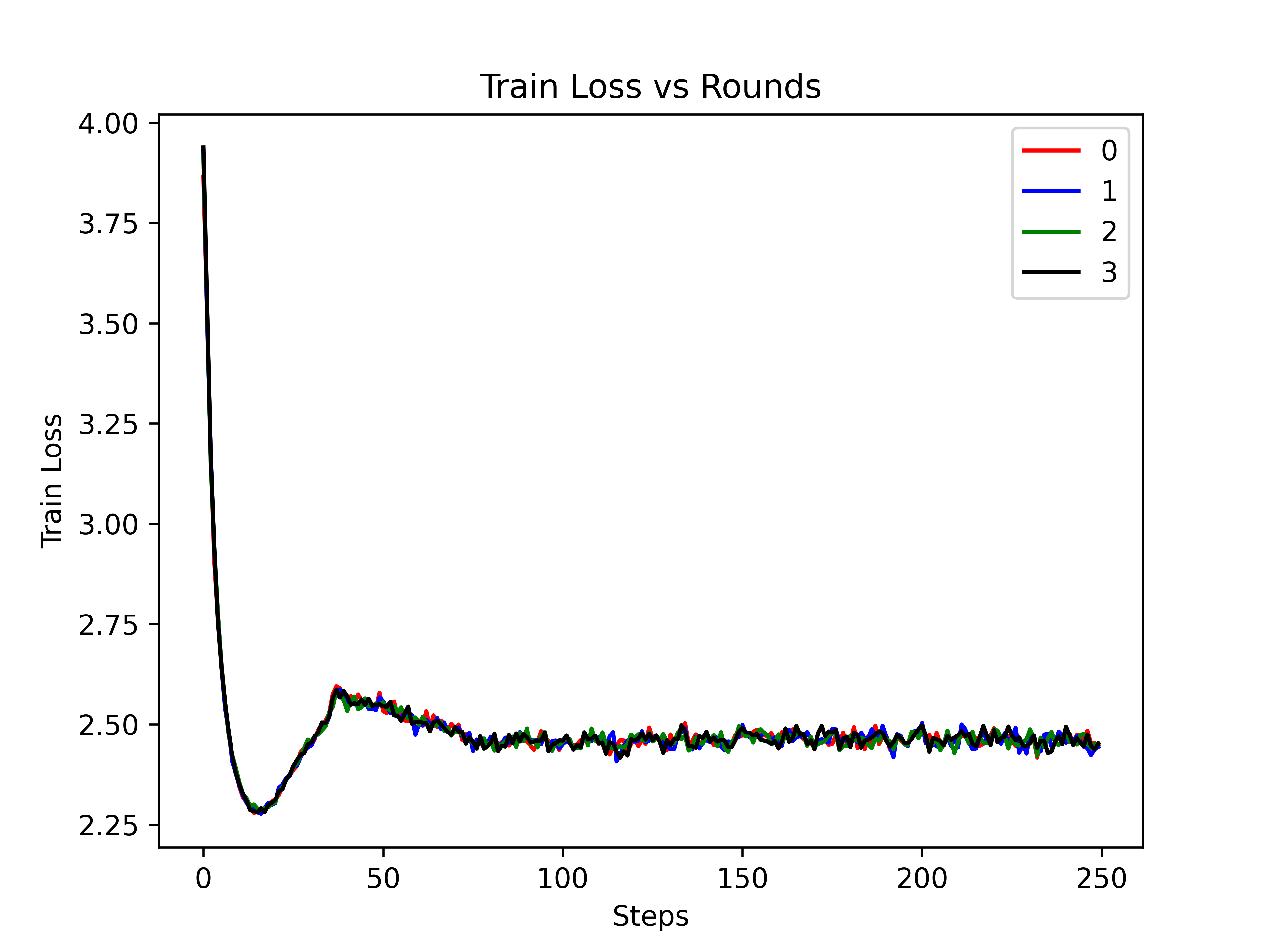}}
    \caption{Federated Effective Rank (FedER) with AdamP and no StepLR}
    \label{fig:feder-adamp-nosteplr}
\end{figure}

Similar to \autoref{fig:fedavg-adamp-nosteplr}, the FedER method demonstrates similar qualities, but wile the results are similar, the FedER underperforms by \textcolor{red}{-1.46\%}. Further investigation is required into how AdamP can improve FedAvg \& FedER and thus should improve with the recommendations outlined in \autoref{fig:fedavg-adam-steplr} such as lowering the learning rate and lowering the weight-decay parameter. 

\newpage
\subsubsection{AdamP with StepLR}

\begin{figure}[H]
    \centering
    \subfigure[Accuracy vs. Steps]{\includegraphics[width=0.5\textwidth]{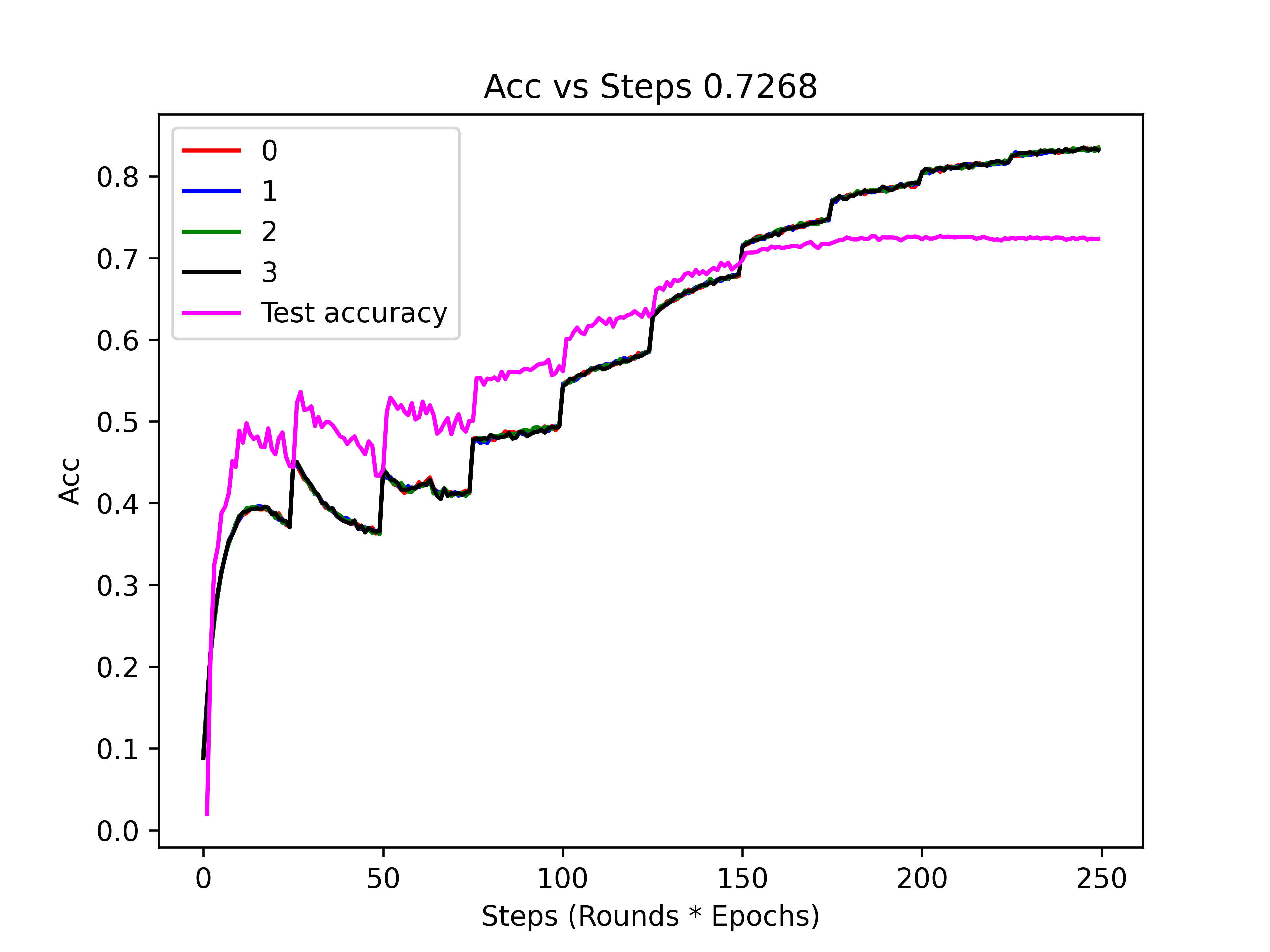}}\hspace{-0.80cm}
    \subfigure[Loss vs Steps]{\includegraphics[width=0.5\textwidth]{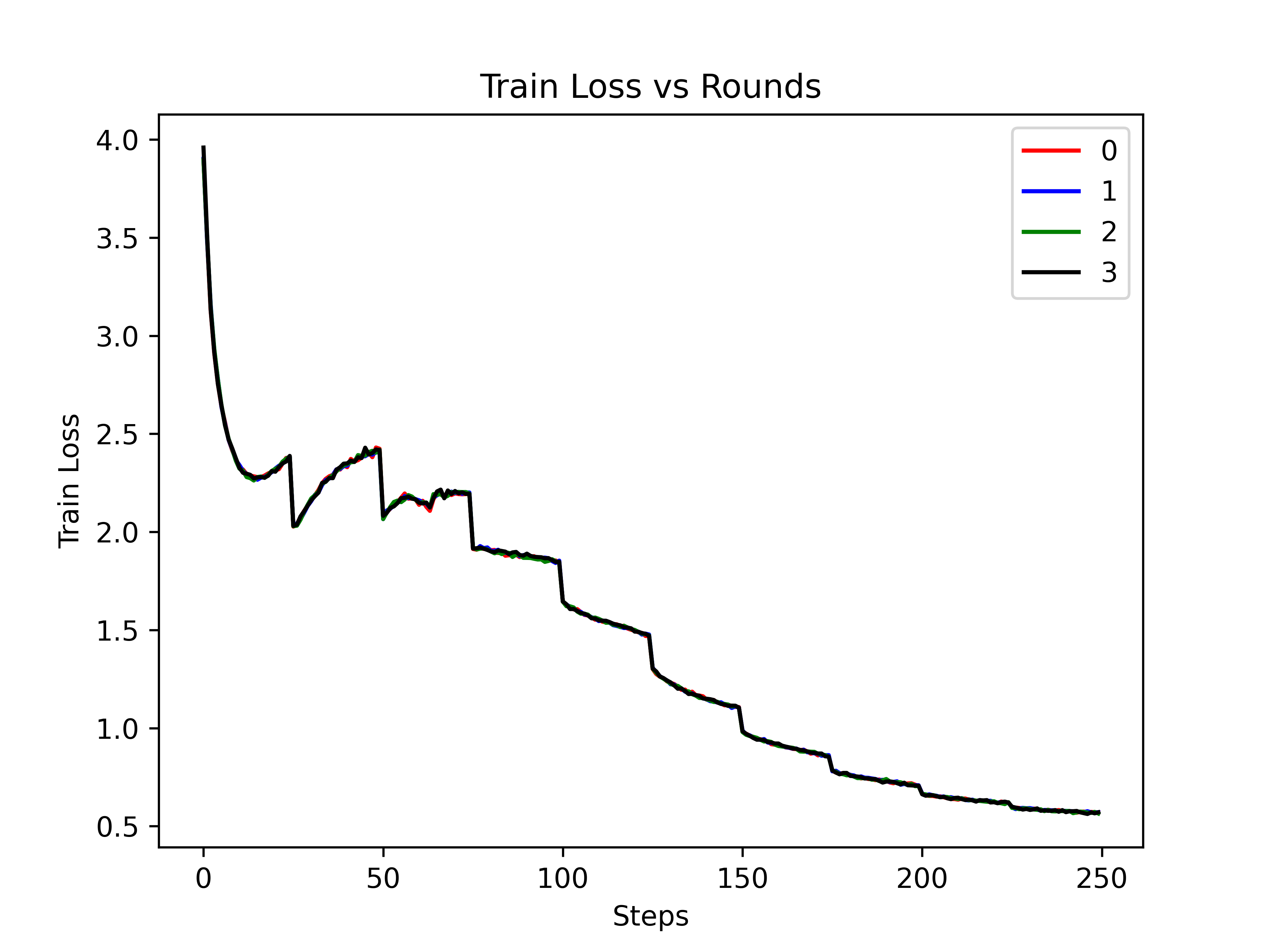}}
    \caption{Federated Effective Rank (FedER) with AdamP and StepLR}
    \label{fig:feder-adamp-steplr}
\end{figure}

Once again, the FedER method performs similarly to FedAvg with AdamP \& StepLR, but this method outperforms by \textcolor{ao}{+0.83\%}. 

\newpage
\subsubsection{RMSGD with no StepLR}
\begin{figure}[H]
    \centering
    \subfigure[Accuracy vs. Steps]{\includegraphics[width=0.5\textwidth]{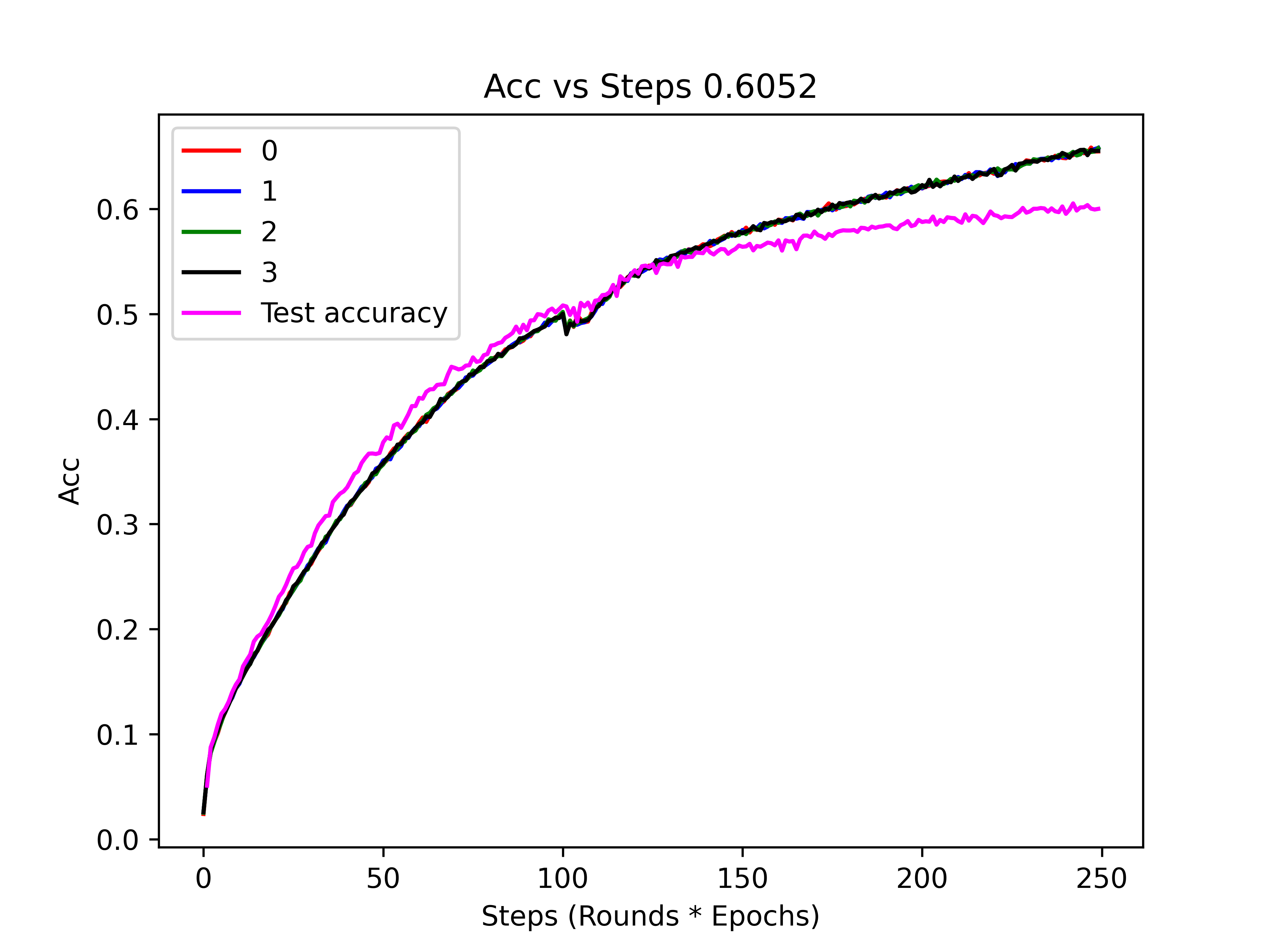}}\hspace{-0.80cm}
    \subfigure[Loss vs Steps]{\includegraphics[width=0.5\textwidth]{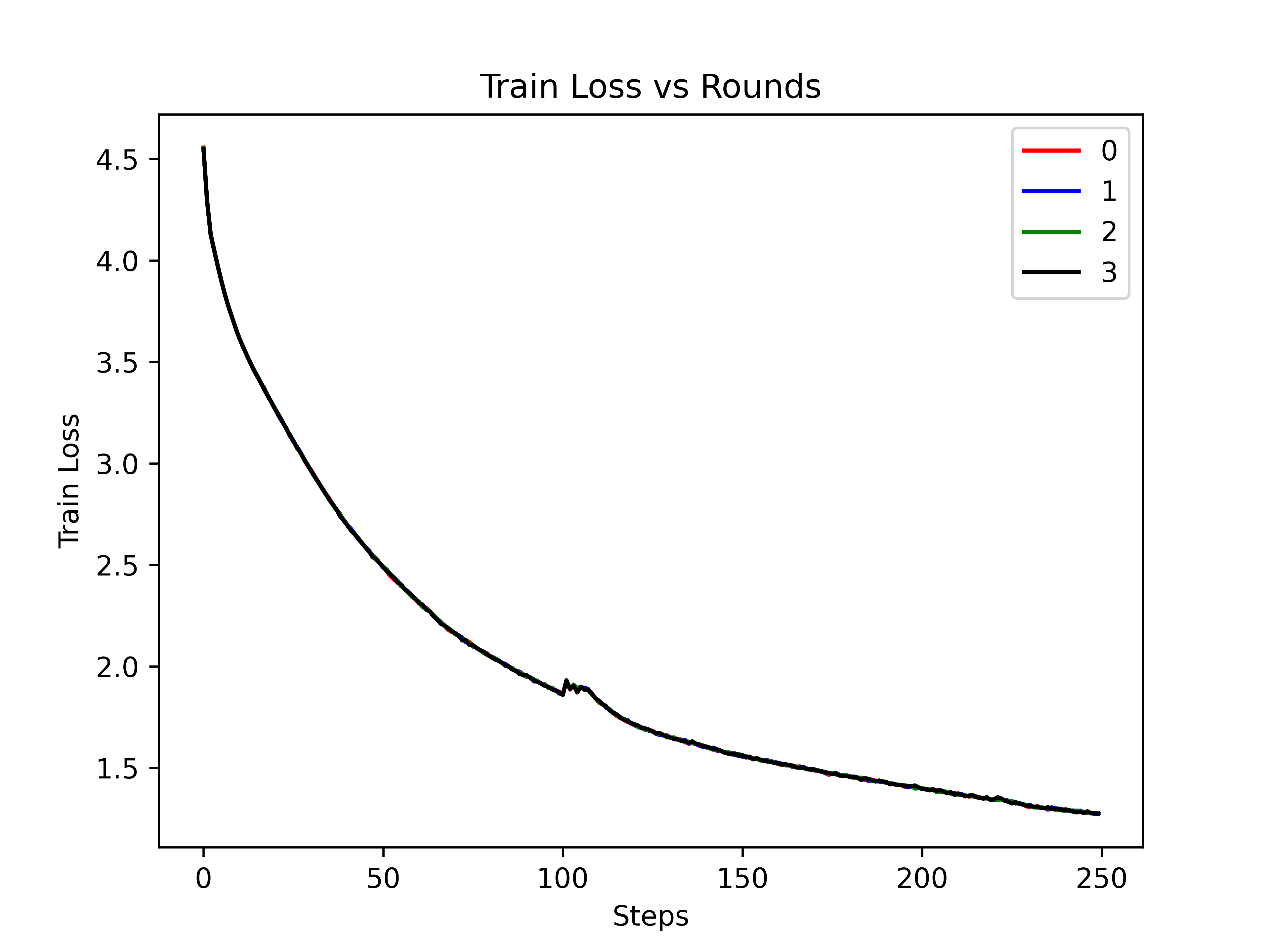}}
    \caption{Federated Effective Rank (FedER) with AdamP and StepLR}
    \label{fig:feder-rmsgd-nosteplr}
\end{figure}












\section{Final Results}

\begin{table}[H]
\begin{center}
\begin{tabular}{c|c|c|c|l}
\textbf{Configuration}                                                & \textbf{\begin{tabular}[c]{@{}c@{}}Non-Fed\\ Baseline\end{tabular}} & \textbf{FedAvg}               & \textbf{FedER}                & \textbf{Difference}                                  \\ \hline
\textit{\begin{tabular}[c]{@{}c@{}}Adam w/out\\ StepLR\end{tabular}}  & 71.71                                                               & 76.87                         & \cellcolor[HTML]{9AFF99}77.05 & {\color[HTML]{009901} +0.18}                         \\ \hline
\textit{\begin{tabular}[c]{@{}c@{}}Adam with \\ StepLR\end{tabular}}  & 74.54                                                               & 75.71                         & \cellcolor[HTML]{9AFF99}75.97 & {\color[HTML]{009901} +0.26}                         \\ \hline
\textit{\begin{tabular}[c]{@{}c@{}}AdamP w/out\\ StepLR\end{tabular}} & \cellcolor[HTML]{FFFFFF}70.06                                       & \cellcolor[HTML]{9AFF99}50.89 & 49.43                         & \cellcolor[HTML]{FFFFFF}{\color[HTML]{FE0000} -1.46} \\ \hline
\textit{\begin{tabular}[c]{@{}c@{}}AdamP with\\ StepLR\end{tabular}}  & \cellcolor[HTML]{FFFFFF}76.39                                       & 71.85                         & \cellcolor[HTML]{9AFF99}72.68 & {\color[HTML]{009901} +0.83}                         \\ \hline
\textit{\begin{tabular}[c]{@{}c@{}}RMSGD w/out\\ StepLR\end{tabular}} & 60.22                                                               & 60.07                         & \cellcolor[HTML]{9AFF99}60.52 & {\color[HTML]{009901} +0.45}                        
\end{tabular}
\end{center}
\caption{Final Top-1 Test-Accuracies for all experiments. Green coloured boxes represent the `better' federated method, and the differences display by how much. }\label{tab:finalresults}
\end{table}

Above in \autoref{tab:finalresults}, the final results are displayed. While the Federated Effective Rank method under-performs quite a lot using AdamP and no StepLR, it out-performs in every other configuration. While investigating the under performance of FedER using AdamP without StepLR, it showed that heavy regularization was occurring and thus, the model was unable to learn effectively. This is supported by the fact that in both FedAvg and FedER settings, the train-accuracy is never able to overtake the test-accuracy, demonstrating that the learned parameters were having trouble generalizing. When comparing FedER \& FedAvg in AdamP using StepLR, we can see that FedER then out-performs FedAvg by 0.83\%, and in this case the StepLR was attempting to minimize oscillations in the loss-landscape, which has the auxiliary value of attempting to over-fit the on the training data. This competition between over-fitting and regularization is part of the reason the model actually learns.\\

Further analysis into investigating the nature of Federated Effective Rank and how other learning metrics like the stable rank from RMSGD \cite{hosseini2022explainable} can be leveraged in order to improve knowledge aggregation in federated settings.

    \chapter{Conclusion \& Future Work}
This work has investigated using novel `learning metrics' to improve Federated Learning, in particular using Effective Rank as the conduit for a novel aggregation method. This work has \textbf{(1)} given the first `learning metrics' aggregation method in a federated setting \textbf{(2)} shown that these `learning metrics', particularly effective rank, is well-suited for this class of problems and \textbf{(3)} developed a novel weight-aggregation scheme relying on these metrics. Therefore, this thesis is concluded by highlighting some open problems and interesting directions for future work. 

\section{Future Work}\label{sec:future-work}
\textbf{Utilizing Stable Rank:} As proposed in \cite{hosseini2022explainable}, stable rank is another novel learning metric that can be used to derive useful results. Furthermore, by utilizing the proof of monotonic increase in stable rank as a learning rate selector, an extension can be proposed to show that a monotonic increase in stable rank across models could lead to improved model aggregations. This is directly related to the next section where convergence of weighted averages of any metric should be proven.\\

\textbf{Proof of convergence:} A proof of convergence must be proposed in order to demonstrate that the effective rank (or any learning metrics) as a weighted average will converge. The basis for this work has already been undertaken in \cite{karimireddy2020scaffold}. \\

\textbf{Leveraging Peer-to-Peer knowledge distillation:} As proposed in \cite{sodhani2020closer}, a peer-to-peer knowledge aggregation scheme can be introduced using co-distillation. Leveraging co-distillation as a method of knowledge aggregation, in combination with learning metrics is an area that is currently unexplored and could pose interesting challenges.\\

\textbf{Non-Identical Models:} As demonstrated in this work, when identical ResNet18 models are trained, they are able to share knowledge seamlessly though the Federated Effective Rank scheme. This scheme breaks down in the case of non-identical models as the layer-by-layer effective rank weighting no longer holds. Consequently, using \autoref{eqn:modeler} would allow non-identical models to leverage effective rank, but at a decreased resolution with respect to the individual weights. Further work into how to aggregate learning metrics per-model while maintaining layer-resolution is an interesting area of research. Additional investigation into a method of analyzing where similar feature representations are being generated and inserting learning metrics at those locations could be a valuable avenue of research.
  \appendix
    \chapter{Utilizing Stable Rank in a Federated Setting}
Regarding Future Work in \autoref{sec:future-work}, initial investigations into leveraging stable rank into the federated setting has been conducted for Adam without StepLR. 

\begin{figure}[H]
    \centering
    \subfigure[Accuracy vs. Epochs]{\includegraphics[width=0.5\textwidth]{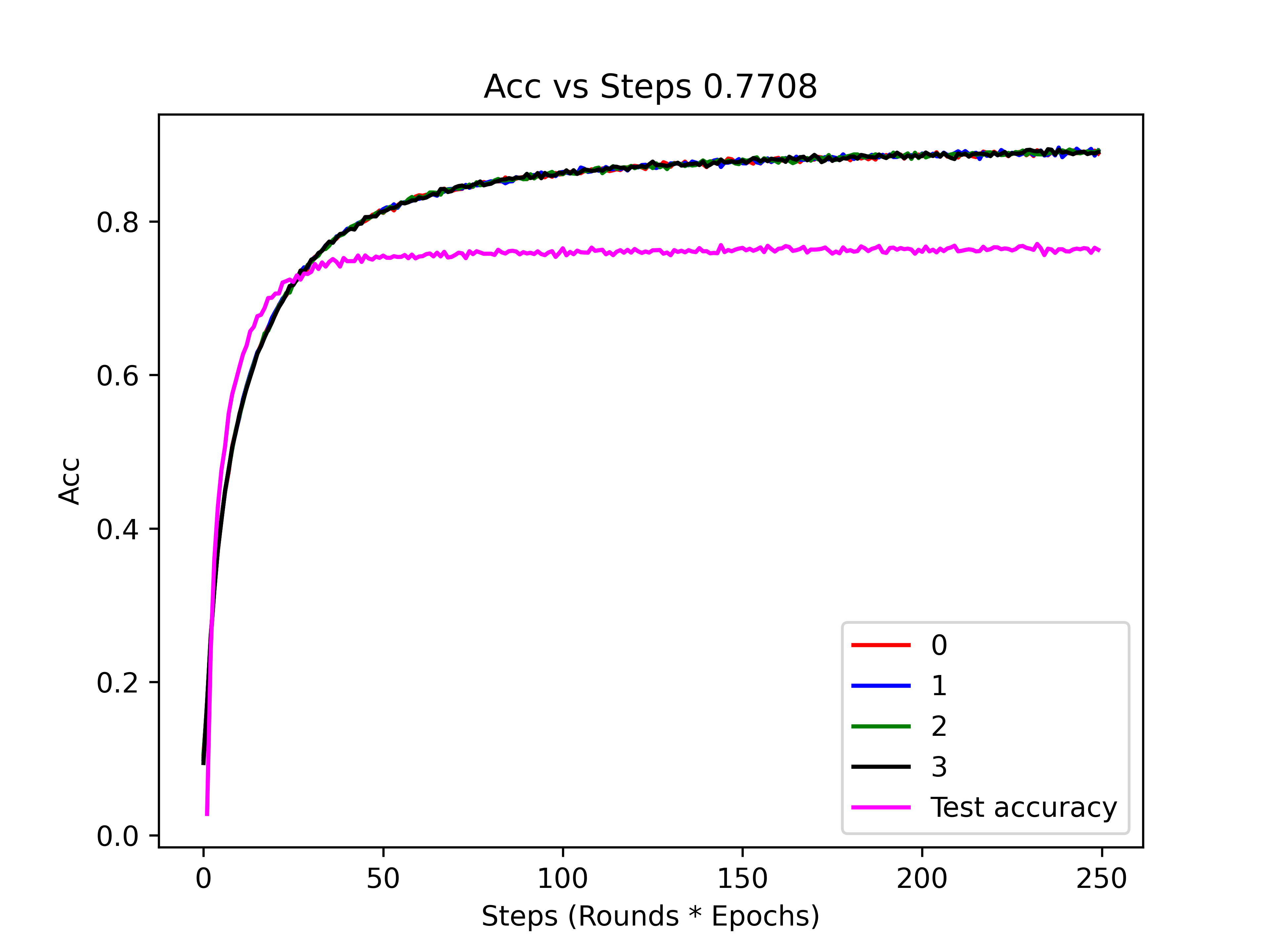}}\hspace{-0.80cm}
    \subfigure[Loss vs Epochs]{\includegraphics[width=0.5\textwidth]{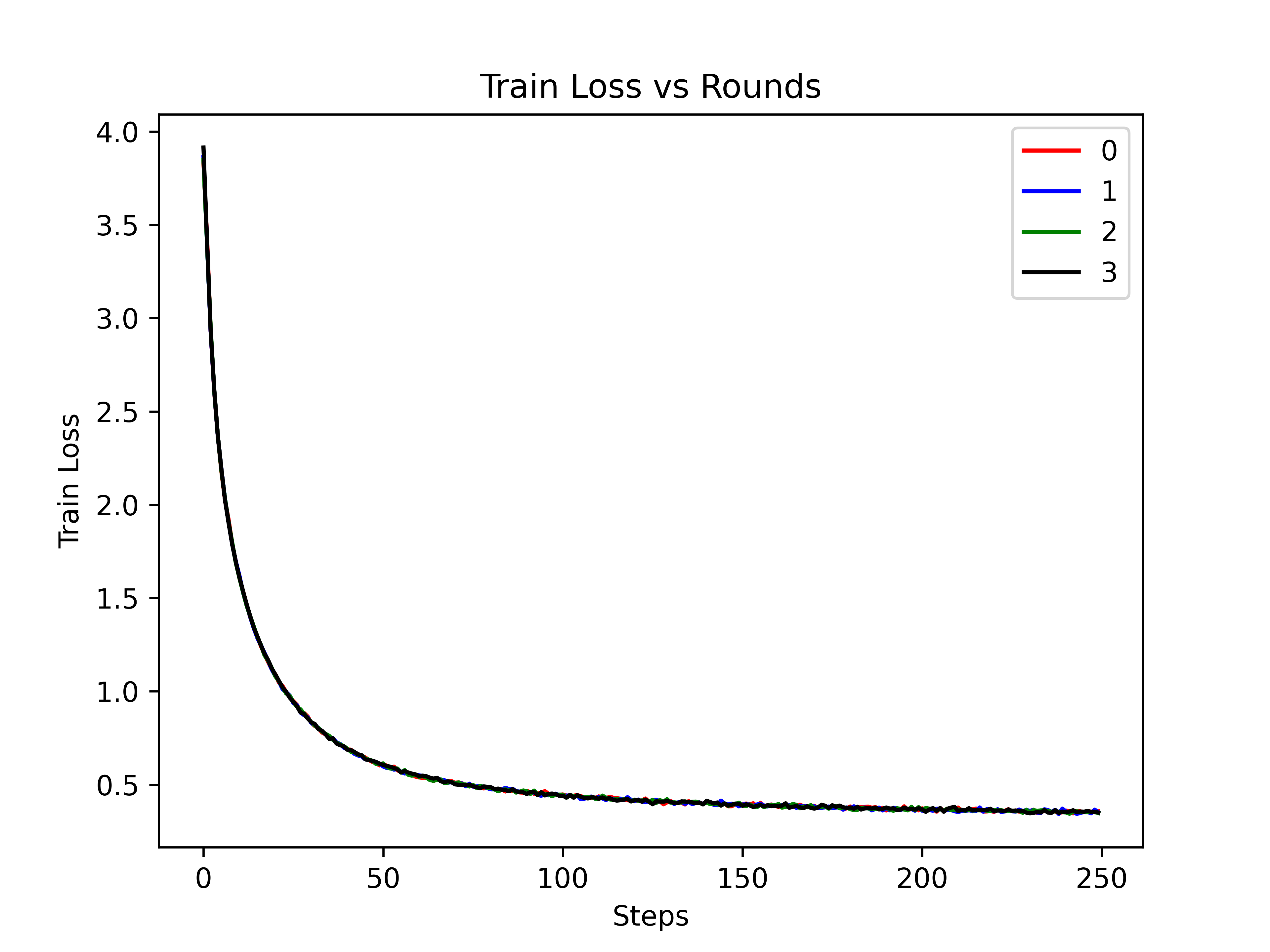}}
    \caption{Federated Stable Rank using Adam without StepLR}
    \label{fig:fedsr-adam-nosteplr}
\end{figure}

As seen in \autoref{fig:fedsr-adam-nosteplr}, the test-accuracy is competitive using stable-rank, and should warrant further investigation.
  \backmatter
  \printbibliography[heading=bibintoc]
\end{document}